\documentclass[11pt, a4paper, logo, twocolumn, copyright]{googledeepmind}

\makeatletter
\renewcommand\bibentry[1]{\nocite{#1}{\frenchspacing\@nameuse{BR@r@#1\@extra@b@citeb}}}
\makeatother

\usepackage{kantlipsum, lipsum}
\usepackage{dsfont}
\usepackage{dm-colors}

\usepackage{algorithm}
\usepackage[noend]{algpseudocode}
\usepackage{adjustbox}
\usepackage{multirow}
\usepackage{nicefrac}

\usepackage{amsthm}
\newtheorem*{remark}{Remark}
\usepackage{tcolorbox}

\definecolor{boxblue}{HTML}{c6dbfc}
\definecolor{boxyellow}{HTML}{ffe5b2}
\definecolor{boxred}{HTML}{f3c0bd}

\usepackage{xcolor} % Make sure xcolor is included

\usepackage[dvipsnames]{xcolor}% http://ctan.org/pkg/xcolor

\makeatletter
\newcommand{\algcolor}[2]{%
  \hskip-\ALG@thistlm\colorbox{#1}{\parbox{\dimexpr\linewidth-2\fboxsep}{\hskip\ALG@thistlm\relax #2}}%
}

\makeatother

\newcolumntype{R}[2]{%
    >{\adjustbox{angle=#1,lap=\width-(#2)}\bgroup}%
    l%
    <{\egroup}%
}

\usepackage[authoryear, sort&compress, round]{natbib} 
\usepackage{comment}

\algrenewcommand\algorithmicindent{0.7em}

\algblock{ParFor}{EndParFor}
\algnewcommand\algorithmicparfor{\textbf{parallel for}}
\algnewcommand\algorithmicpardo{\textbf{do}}
\algnewcommand\algorithmicendparfor{\textbf{end\ parallel for}}
\algrenewtext{ParFor}[1]{\algorithmicparfor\ #1\ \algorithmicpardo}
\algrenewtext{EndParFor}{\algorithmicendparfor}

\usepackage{pifont}% http://ctan.org/pkg/pifont

\usepackage{subcaption}
\usepackage{url}

\graphicspath{{figures/}}

\title{{\fontsize{18.7pt}{1pt}\selectfont Streaming DiLoCo with overlapping communication:}\\ {\fontsize{15pt}{1pt}\selectfont Towards a Distributed Free Lunch \includegraphics[scale=0.1]{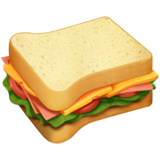}}}

\makeatletter
\def\@headertitle{Streaming DiLoCo with overlapping communication: Towards a Distributed Free Lunch}
\makeatother

\correspondingauthor{douillard@google.com}

\reportnumber{} % Leave blank if n/a 

\renewcommand{\today}{} 

\author[*,1]{Arthur Douillard}
\author[*,1]{Yanislav Donchev}
\author[2]{Keith Rush}
\author[$\dagger$,2]{Satyen Kale}
\author[2]{Zachary Charles}
\author[2]{Zachary Garrett}
\author[3]{Gabriel Teston}
\author[1]{Dave Lacey}
\author[1]{Ross McIlroy}
\author[1]{Jiajun Shen}
\author[1]{Alexandre Ramé}
\author[1]{Arthur Szlam}
\author[1]{Marc'Aurelio Ranzato}
\author[1]{Paul Barham}

\affil[1]{Google DeepMind}
\affil[2]{Google Research}
\affil[3]{Google}
\affil[*]{Equal core contributions}
\affil[$\dagger$]{Currently at Apple.}

\begin{abstract}
Training of large language models (LLMs) is typically distributed across a large number of accelerators to reduce training time. Since internal states and parameter gradients need to be exchanged at each and every single gradient step, all devices need to be co-located using low-latency high-bandwidth communication links to support the required high volume of exchanged bits. Recently, distributed algorithms like DiLoCo~\citep{douillard2023diloco} have relaxed such co-location constraint: accelerators can be grouped into ``workers'', where synchronizations between workers only occur infrequently.  This in turn means that workers can afford being connected by lower bandwidth communication links without affecting learning quality. However, in these methods, communication across workers still requires the same peak bandwidth as before, as the synchronizations require all parameters to be exchanged across all workers.  In this paper, we improve DiLoCo in three ways.  First, we synchronize only subsets of parameters in sequence, rather than all at once, which greatly reduces peak bandwidth.  Second, we allow workers to continue training while synchronizing, which decreases wall clock time.  Third, we quantize the data exchanged by workers, which further reduces bandwidth across workers.  By properly combining these modifications, we show experimentally that we can distribute training of billion-scale parameters and reach similar quality as before, but reducing required bandwidth by two orders of magnitude.

\end{abstract}

\begin{document}

\maketitle

\section{Introduction} \label{sec:intro}

Scaling deep learning has led to significant leaps in capability \citep{geminiteam2024geminifamilyhighlycapable,openai2024gpt4technicalreport,grattafiori2024llama3herdmodels}. Although neural architectures have evolved over the past decade, the standard approach to optimization remains essentially unchanged from the days of Alexnet \citep{krizhevsky2012alexnet}.  Practitioners use minibatch stochastic gradient descent, with backpropagation through the model's layers to compute the gradients.   As in Alexnet, which already combined two hardware accelerators for parallel training, models are trained with multiple hardware accelerators.

However, modern training runs, for example for large language models (LLM), may use tens of thousands of accelerators, and this number increases year after year.  Building and maintaining a data-center that can co-locate that many accelerators is expensive and leads to increasingly complex engineering challenges.  Beyond the physical infrastructure, orchestrating the passage of gradients, parameters and intermediate states between these devices at each optimization step, while keeping all devices fully utilized is technically challenging from a software engineering perspective. Furthermore, the more devices that are used for each synchronous training step, the more chances there are that one of them fails, risking halting training, or introducing subtle numerical issues.  %mention something about centralization/single entity?

Recent publications \citep{douillard2023diloco,jaghouar2024intellect1}, building on work by \cite{mcmahan2017fedavg},
have demonstrated that the co-location requirements of all accelerators can be loosened.  These methods allow highly-performant training when accelerators are grouped into several ``workers'' with fast bandwidth intra-worker but with slow bandwidth inter-workers. The basic approach is to allow each worker to continue training for many minibatches, independently of the other workers; and then synchronize the parameters of the workers after a set number of these ``inner'' steps.  The synchronization, in its simplest form \citep{mcmahan2017fedavg}, is to average the parameters of the workers \citep{wortsman2022soup}; but more sophisticated methods \citep{huo2020outernesterov, reddi2021adaptive} use the workers' parameters to form a pseudo-gradient to update the shared parameters.  The details of the formulation of the weight synchronization is important for machine-learning efficiency; see Section \ref{sec:model_diloco}.  

However, in these approaches, the synchronization typically requires an all-reduce operation which fully synchronizes the model parameters on some step. This all-reduce results in two main issues: 1) a large peak bandwidth, and 2) a blocking of the workers while they wait to receive updated weights.  In this work, dubbed {\em Streaming DiLoCo}, we propose three modifications to these approaches to practically reduce the peak bandwidth and mitigate worker-blocking without loss of learning efficiency:

\begin{tcolorbox}[colback=boxblue, colframe=black, arc=4pt, boxsep=0.3pt]% 
\textbf{Contribution 1: \emph{Synchronization}.} 

We synchronize subsets of parameters on a schedule, rather than all parameters at once. This contribution reduces the peak required bandwidth.
\end{tcolorbox}%
\begin{tcolorbox}[colback=boxyellow, colframe=black, arc=4pt, boxsep=0.3pt]% 
\textbf{Contribution 2: \emph{Overlapping}.} 

We overlap worker computation and communication of synchronizations. This contribution increases the tolerated latency of communication. 
\end{tcolorbox}%
\begin{tcolorbox}[colback=boxred, colframe=black, arc=4pt, boxsep=0.3pt]% 
\textbf{Contribution 3: \emph{Quantization}.} 

We compress the outer gradients to four bits per parameters without loss of performance. This contribution reduces the total amount of bits exchanged.
\end{tcolorbox}%

We show experimentally that our model, Streaming DiLoCo, is strictly superior to the original DiLoCo \citep{douillard2023diloco}, and achieves similar performance to the bandwidth-costly data-parallelism. Since we attain the same quality at negligible bandwidth, we consider our approach as an important stepping stone towards a form of \textbf{{\em distributed} free lunch}.

\section{Model} \label{sec:model}

For all algorithms, we denote the model parameters as $\theta$. We use the superscript notation $\theta^{(t)}$ to indicate the parameters at a given step $t$, and the subscript notation $\theta_m$ to denote a particular shard of the DiLoCo replica. For example, $\theta^{(t)}_m$ indicates the parameters of DiLoCo replica $m$ at step $t$. If no subscript is used, the parameters are replicated across DiLoCo replicas. Note that it is possible for parameters to not be replicated and yet to be of the same value.

\begin{algorithm}
\caption{FedOpt / DiLoCo} \label{alg:diloco}
\begin{algorithmic}[1]
\Require $M$ replicas
\Require Synchronization frequency $H$
\Require Model replicas $\{\theta^{(t-1)}_1, \dots, \theta^{(t-1)}_M\}$
\Require Data shards $\{\mathcal{D}_1, \dots, \mathcal{D}_M\}$
\Require Optimizers $\texttt{InnerOpt}$ and $\texttt{OuterOpt}$
\ParFor{\texttt{replica $m = 1 \ldots M$}} 
\For{\texttt{step $t = 1 \ldots T$}}
    \State $x \sim \mathcal{D}_m$
    \State $\mathcal{L} \gets f(x, \theta_m^{(t-1)})$
    \State $\theta_m^{(t)} \gets \texttt{InnerOpt}(\theta_m^{(t-1)}, \nabla_\mathcal{L})$
    \item[]
    \If{$t\mod H == 0$}
        \State $\Delta^{(t)}_{m} \gets \theta^{(t-H)}_{m} - \theta_{m}^{(t)}$
        \State $\Delta^{(t)} \gets {\small \texttt{async-send}}[\frac{1}{M} \sum_{m=1}^M (\Delta^{(t)}_{m})]$
        \State $\texttt{block-receive}[{\Delta^{(t)}}]$
        \State $\theta_m^{(t)} \gets \texttt{OuterOpt}(\theta_m^{(t-H)}, \Delta^{(t)})$
    \EndIf
\EndFor
\EndParFor
\end{algorithmic}
\end{algorithm}

\subsection{Context: FedOpt and DiLoCo} \label{sec:model_diloco}
%\mr{Update the line numbers in the algorithm when finalized.}
FedOpt~\citep{reddi2021adaptive} is a generic framework to perform federated learning with a bi-level optimization. $M$ local replicas perform $H$ steps of \textit{inner} independent optimizations on a different subset of the data (L3 to L5 in \autoref{alg:diloco}). Every $H$ steps, each replica computes an \textit{outer gradient} $\Delta_m^t = \theta_m^{(t-H)} - \theta_m^{(t)}$ (L7), a delta in the parameters space, and communicates to all other replicas. This communication can be performed through a central parameter server or through direct communication of each worker to the others (e.g. with a ring all-reduce), and results in each worker obtaining $\Delta^t = \nicefrac{1}{M} \sum_{m=1}^M \Delta^t_m$ (L7-9). This outer gradient is applied on a set of \textit{outer parameters}, the previously synchronized parameters $\theta_m^{(t-H)}$, with an \textit{outer optimizer} (L10). The full algorithm is shown in \autoref{alg:diloco}.

The costly communication between non-colocated devices happens during the averaging of outer gradients, in L8-9 of \autoref{alg:diloco}. It is as costly as in Data-Parallel, but instead of being executed at every step, it is done every $H$ (e.g. one hundred) steps, thus amortizing the communication cost.

\begin{figure*}[ht!]
    \centering
    \includegraphics[width=0.9\linewidth,trim={0cm 0cm 5cm 7.5cm},clip]{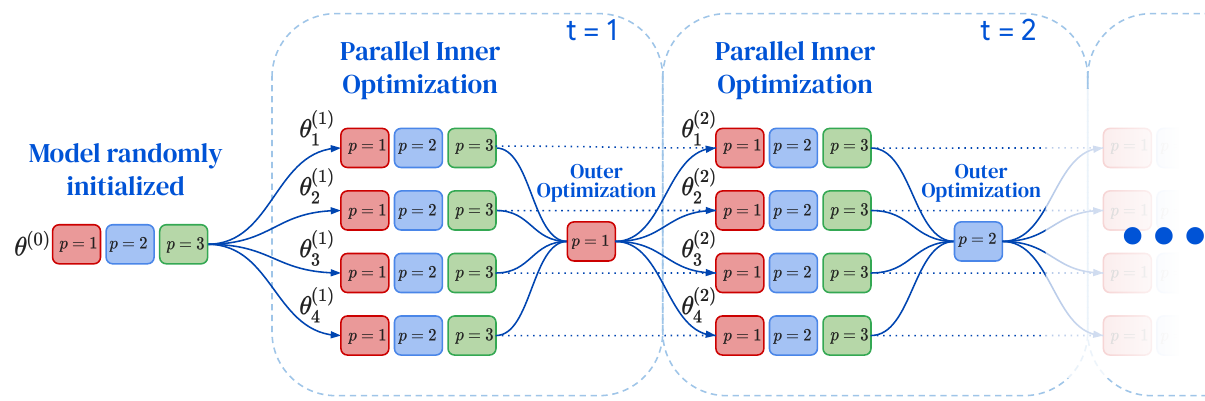}
    \caption{\textbf{Streaming DiLoCo}: each replica trains independently for dozen of inner optimization steps, and then synchronize a single fragment during outer optimization. In this figure, there are $M=4$ replicas with $p=\{1,2,3\}$ fragments. Each fragment can be made of several transformer layers. Note that this figure only showcases the streaming partial updates (\autoref{sec:model_streaming}) and not the quantized communication overlapping (subsection \ref{sec:model_overlapping} and \ref{sec:model_low_precision}).}
\label{fig:streaming_diloco}
\end{figure*}

DiLoCo is a successful instantiation of FedOpt applied to language models where the inner optimizer is Adam \citep{kingma2014adam} and the outer optimizer is SGD with Nesterov momentum \citep{sutskever2013nesterov}. 
In this work, which focuses on distributed optimization, and unlike in the federated learning literature (as discussed in FedOpt), the workers aren't sampled; but instead all workers will be present at each step.

\subsection{Streaming partial updates} \label{sec:model_streaming}

\begin{figure}[ht!]
    \centering
    \includegraphics[width=0.8\linewidth,trim={0cm 0cm 16cm 0.75cm},clip]{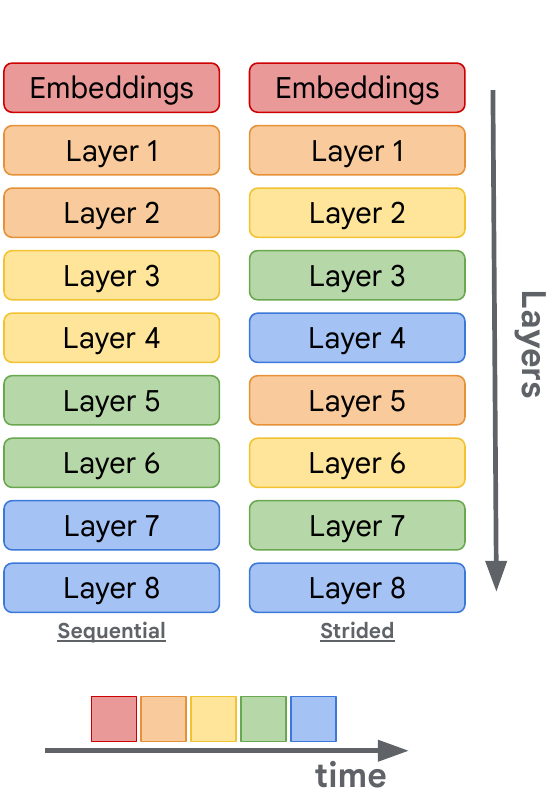}
    \caption{\textbf{Streaming pattern}: sequential (left) and strided (right). Colors denotes the fragment. A different fragment is synchronized each time.}
\label{fig:streaming_pattern}
\end{figure} 

Instead of communicating the full outer gradient vector ($\Delta_m^{(t)}, \forall m \in \{1, ... M\}$) every $H$ steps, we propose to share only a fragment $p$ of it ($\Delta_{m,p}^{(t)}$) more frequently, as highlighted in \autoref{fig:streaming_diloco}.  There is a huge possible space of choices for these fragments and specification of ``more frequently''\footnote{For example, an extreme version might be to send a constant bitstream of random choices (according to some optimization-useful distribution) of parameters.};  here we consider the simple partition of our network into $P$ fragments made of several transformer blocks.  Specifically, we study two fragment patterns, as shown in \autoref{fig:streaming_pattern}: 1) sequential where each fragment comprises consecutive transformer blocks and 2) strided where each fragment is composed of interleaved transformer blocks. We will demonstrate in \autoref{sec:ablations} that the algorithm is robust to the particular choice of fragments. Since the stride version offers in practice slightly better compute utilization (less time spent communicating instead of computing), we will use it as the default choice in our experiments. As we increase model scale, the fragment definition (e.g., how many transformer blocks comprise a fragment) is maintained, which means that larger models have more fragments.

The resulting algorithm in shown is \autoref{alg:streaming} (and contrasted with the original version shown in \autoref{alg:diloco}), where only a fragment $p$ of the replica $m$ is shared. We denote a fragment with a new lower script, thus $\theta^{(t)}_{m,p}$ is the parameters of fragment $p$  of the replica $m$ at step time $t$.

The Streaming DiLoCo's inner optimization (L3-5 of \autoref{alg:streaming}) is identical to DiLoCo's (L3-5 of \autoref{alg:diloco}). However, the outer optimization (L12) is done per fragment.  If a fragment $p$ satisfies the condition $t+t_p \mod H = 0$, where $t_p$ is a time offset fragment-dependent, then it is synchronized.  In this way, each fragment will always do $H$ steps before being synchronized, but overall the model is synchronizing \textit{some} fragment more frequently than every $H$ steps. For example, with $H=100$ and $P=2$ fragments, the first fragment will be synchronized at step $t=100$, $t=200$, ... ($t_{p=1} = 0$); the second fragment will be synchronized at step $t=150$, $t=250$, ... ($t_{p=2} = 50$). While in practice, given an equal $H$, streaming DiLoCo communicates more often than DiLoCo, the peak communication is reduced by a factor of $\nicefrac{|p|}{L}$ with $|p|$ the size of a fragment in \textit{layers} and $L$ the total number of layers.
%The embedding is a special case, in its own fragment.
\begin{algorithm}[t!]
\caption{Streaming DiloCo} \label{alg:streaming}
\begin{algorithmic}[1]
\Require $M$ replicas
\Require Number of inner steps $H$
\Require Fragments $p\,\in \{1, \dots, P\}$ with their respective synchronization offset $t_p$
\Require Model replicas $\{\theta^{(t)}_1, \dots, \theta^{(t)}_M\}$
\Require Inner overlap delay $\tau < H$
\Require Data shards $\{\mathcal{D}_1, \dots, \mathcal{D}_M\}$
\Require Optimizers $\texttt{InnerOpt}$ and $\texttt{OuterOpt}$
\ParFor{\texttt{replica $m = 1 \ldots M$}} 
\For{\texttt{step $t = 1 \ldots T$}}
    \State $x \sim \mathcal{D}_m$
    \State $\mathcal{L} \gets f(x, \theta_m^{(t-1)})$
    \State $\theta_m^{(t)} \gets \texttt{InnerOpt}(\theta_m^{(t-1)}, \nabla_\mathcal{L})$
    \item[]
    %\For {fragment $p = 1, \dots, P$}
        \If{$\exists p$ s.t. $t - t_p \mod H == 0$}
            \State $\Delta^{(t)}_{m,p} \gets \theta^{(t-H)}_{m, p} - \theta_{m,p}$
            \State $\Delta^{(t)}_p \gets {\small \texttt{async-send}}[\frac{1}{M} \sum_{m=1}^M(\Delta^{(t)}_{m,p})]$
            %\State  $\texttt{send}[{\Delta^{(t)}_{p}]$
        \EndIf
        \item[]
        \If{$\exists p$ s.t. $t-t_p-\tau \mod H == 0$}
            \State $\texttt{block-receive}[{\Delta^{(t-\tau)}_{p}}]$
            \State $\tilde{\theta}^{(t)}_{m,p} \gets \texttt{\scriptsize OuterOpt}(\theta^{(t-\tau-H)}_{m,p}, \Delta^{(t-\tau)}_{p})$
            \State $\theta_{m,p}^{(t)} \gets \alpha \theta^{(t)}_{m,p} + (1-\alpha)\tilde{\theta}^{(t)}_{m,p}$
        \EndIf
\EndFor
\EndParFor
\end{algorithmic}
\end{algorithm}
\subsection{Overlapping communication with computation} \label{sec:model_overlapping}

To further maximize the time spent doing computation v.s. communication, we propose to overlap the communication of the outer gradient fragment with the inner optimization computation; the overlap happens with a strictly positive $\tau$ in \autoref{alg:streaming}, lines 10-12. At the beginning of outer step $t+1$, instead of waiting for the communication of the fragment $\texttt{block-receive}$,  we immediately start the new round of optimization. After $\tau-1$ inner steps (L3-5), we block-wait for the exchanged fragment (L10), apply the outer optimizer on the previously synchronized fragment ($\theta^{(t-\tau-H)}_{m,p^{(t-\tau)}}$), and merge it with the currently optimized fragment with a mixing factor $\alpha$. $\alpha=1$ is equivalent to no communication between replicas, $\alpha=0$ discards any updates done in the first $\tau$ steps on the fragment $p$, and $\alpha=0.5$ does a uniform average between the local fragment parameters and the globally shared one.

%Notice that only in the newly synchronized fragment needs communication, and thus the communication of L11 can be overlapped with the inner optimization computation of L10-13. When the outer optimizer is as simple as Nesterov momentum, the outer optimization (L21) is practically instantaneous, and could be overlapped by performing its computation on CPU. 
%\zg{Wdyt about an extra sentence to explain that the outer optimizer could be overlapped as well, nothing fundamentally blocking it, esp. in cases where it might actually be expensive? There could be a connection here to LookAhead Optimizer when $M=1$, where this would be an approach that parallelizes more of computation}

\subsection{Low-precision outer gradients} \label{sec:model_low_precision}
The previous methods, streaming and overlapping communication with computation, reduce peak bandwidth and wall-clock time, respectively. To reduce total amount of bits exchanged, we use lower-precision in the outer gradients exchanged by workers (while still using FP32 for computing gradients), up to a float with 4 bits (1 sign bit, 3 exponent bits, and 0 mantissa bit) called \texttt{E3M0} \citep{agrawal2024exmy}. Across a wide variety of experiments, we found  no sign of performance regression when employing such low precision numbers during communication, even at the billion scale. This compression is applied when sending each replica's unreduced outer gradients to miminize the amount bits communicated (L8 of \autoref{alg:streaming}), but once received by a replica, importantly, the accumulation is done in FP32 for stability.

\subsection{Discussion on the memory overhead} \label{sec:model_memory}

In an SPMD\footnote{\url{https://en.wikipedia.org/wiki/Single_program,\_multiple_data}} model, the memory overhead of the Data-Parallel baseline is the parameters ($1\times$) + Adam state ($2\times$). (Streaming or not) DiLoCo's memory overhead is the parameters ($1\times$), the Adam state ($2\times$), the outer global parameters ($1\times$), and the outer Nesterov state ($1\times$).
Thus, our method requires $66\%$ (\nicefrac{5}{3}) more memory compared to Data-Parallel. However, in the case of Streaming DiLoCo, only a \textit{subset} of the outer parameters and outer optimizer state is needed at a given time. Therefore, this overhead can be alleviated by offloading the additional bits onto CPU memory \citep{beaumont2022weightoffloading}. The memory overhead to hold in HBM at any point in time is the size of a fragment $|p|$ times two, for the outer parameters and outer optimizer state. 
For a 100 billion parameter model for instance, with $|p| = $ three layers and with a total of $108$ layers, that amounts to a $2\%$ increase of memory (additional 20 GB to 1,117 GB\footnote{The size in GB of the parameters \& inner Adam optimizer state is the number of parameters $\times$ 3 $\times$ 4 (FP32). The size in GB of the additional fragment and its outer Nesterov optimizer state is the number of parameters $\times$ 2 $\times$ 4 $\times \frac{3}{108}$.}). This extra memory is used when there are no activations or gradients in live memory, and thus should fit in HBM without any problem. 

Furthermore, the communication schedule is deterministic and known prior to training. Thus, we can start the transfer from RAM to HBM of a fragment (and its associated outer optimizer state) while finishing the previous (inner) gradients passes. Given that only a small subset is required at a given time, the memory transfer cost is negligible. With an H100 with PCIe\footnote{\url{https://www.nvidia.com/en-gb/data-center/h100/}}, characterized by 2 TB/s of bandwidth speed, and without any sharding, this transfer is done in less than 10 milliseconds.

\section{Experiments} \label{sec:exps}
We run experiments demonstrating the compute utilization benefits of our approach in a a bandwidth and compute simulation in Section \ref{sec:model_compute_utilization}.  In Section \ref{sec:llm_exps} we show in practice the model learning outcomes. Finally, in Section \ref{sec:ablations} we show results with variations of the three main contributions of the paper, ablating their relative importance.

\subsection{Compute utilization} \label{sec:model_compute_utilization}

\begin{figure*}
\centering
 \includegraphics[width=.99\linewidth]{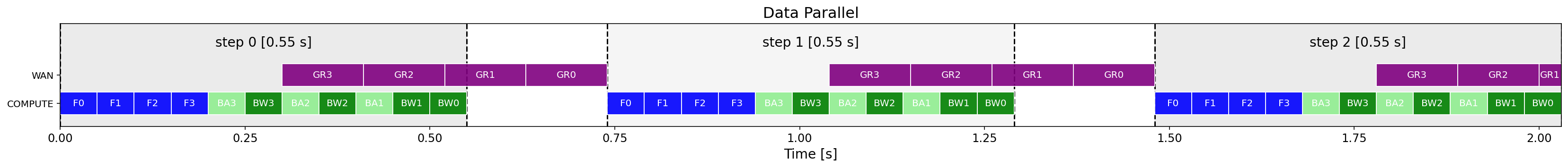}\\
 \includegraphics[width=.99\linewidth]{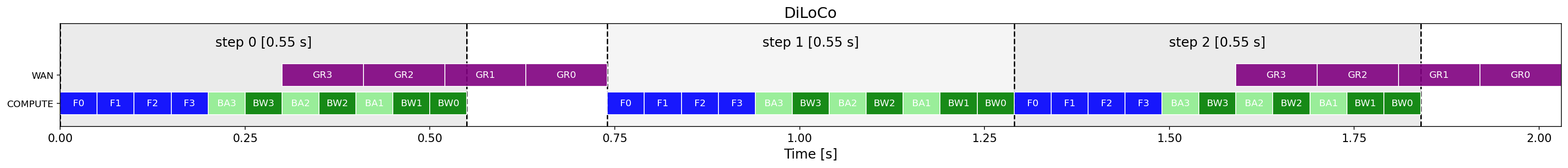} \\
  \includegraphics[width=.99\linewidth]{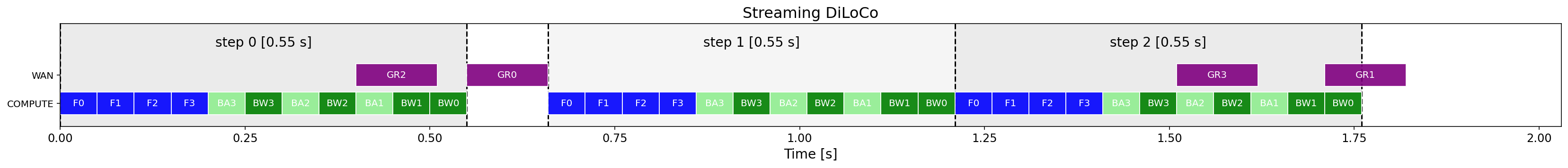} \\
 \includegraphics[width=.99\linewidth]{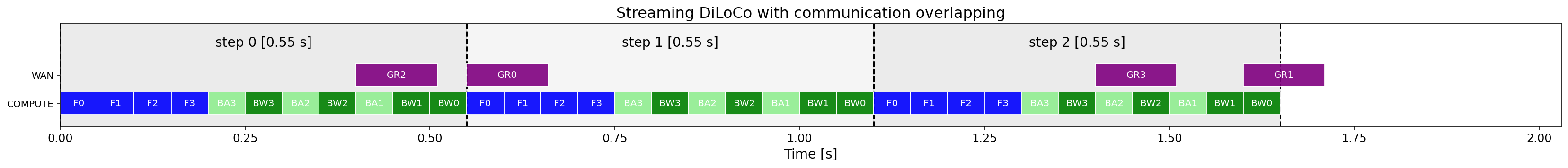} \\
\caption{Simulation of a schedule interleaving forward passes (in \textcolor{blue}{blue}), backward passes w.r.t. activations and parameters (resp. in \textcolor{green}{light} and \textcolor{teal}{dark green}), and (outer) gradient reduction (in \textcolor{violet}{purple}).}
\label{fig:schedule}
\end{figure*}

\begin{figure*}[t]
\centering
\captionsetup[subfigure]{justification=centering}
\begin{subfigure}{0.325\linewidth}
  \centering
  \includegraphics[width=1\linewidth]{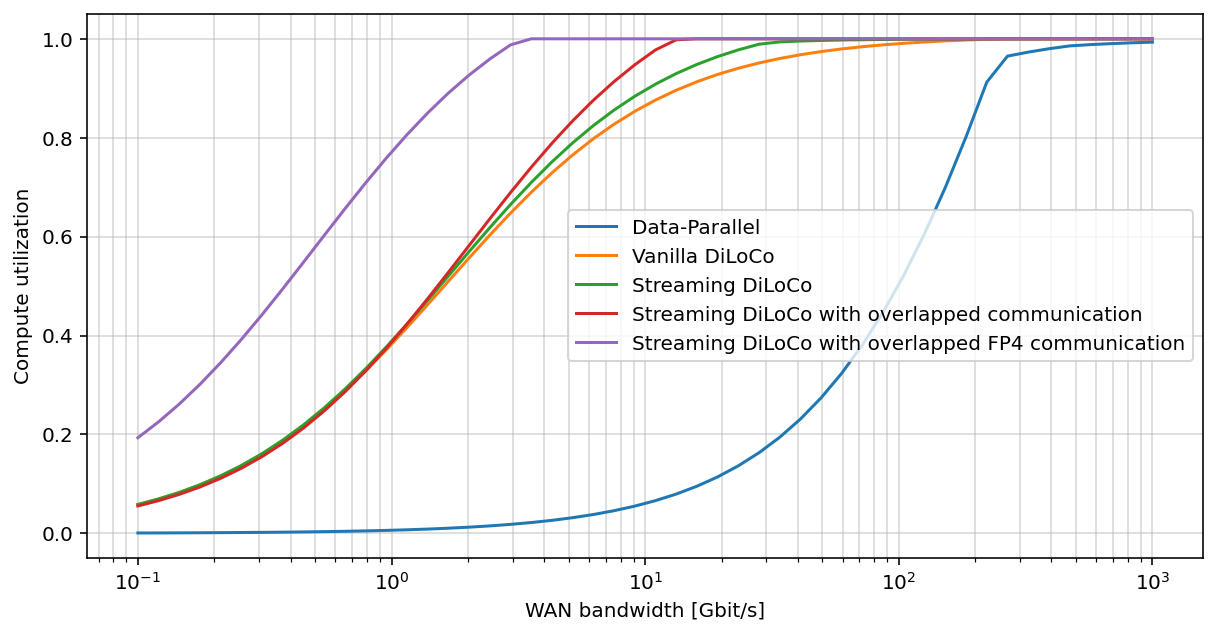}
  \caption{1B parameters model.}
  \label{fig:bandwdith_1b}
\end{subfigure}\hfill
\begin{subfigure}{0.325\linewidth}
  \centering
  \includegraphics[width=1\linewidth]{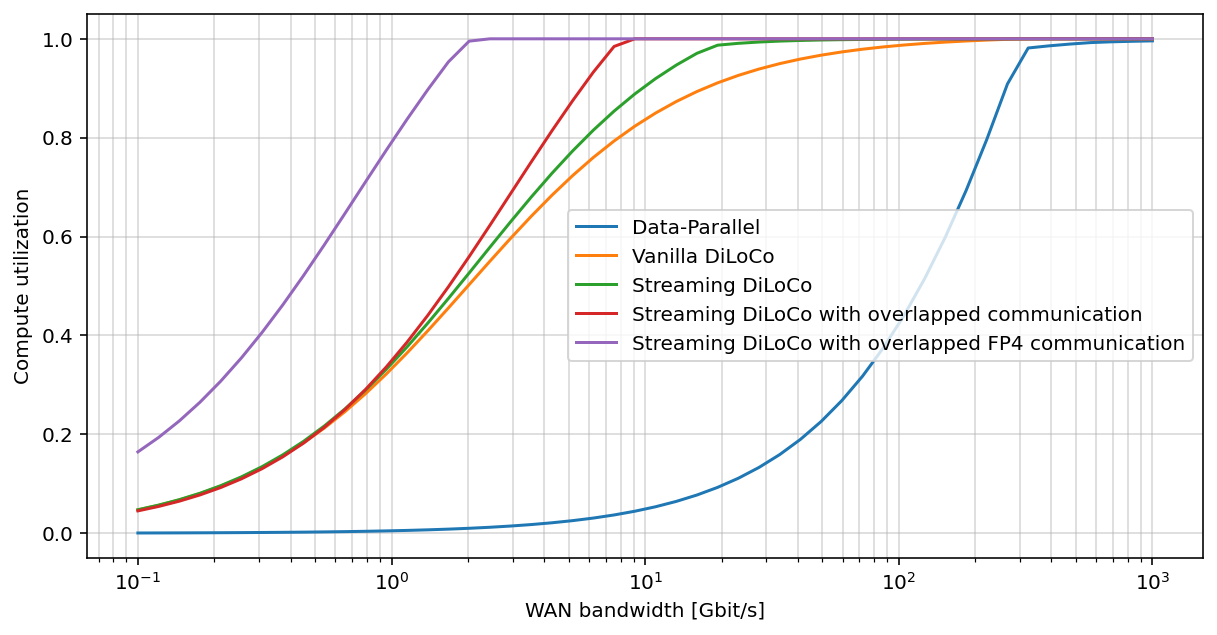}
  \caption{10B parameters model}
  \label{fig:bandwdith_10b}
\end{subfigure}\hfill
\begin{subfigure}{0.325\linewidth}
  \centering
  \includegraphics[width=1\linewidth]{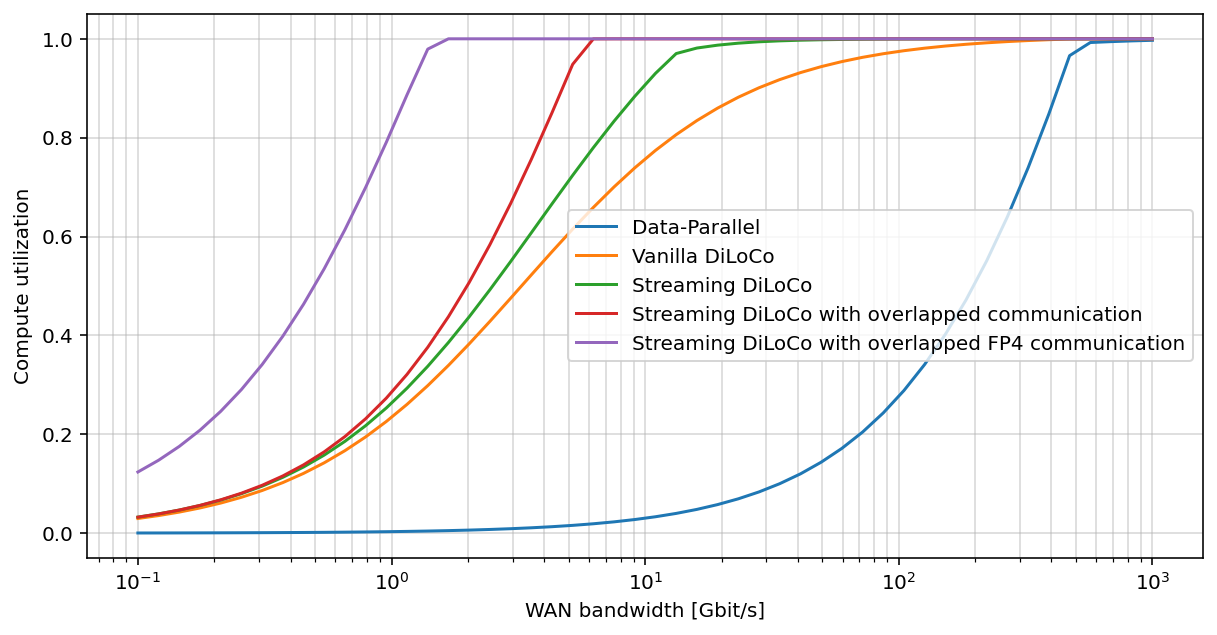}
  \caption{100B parameters model}
  \label{fig:bandwdith_100b}
\end{subfigure}
\caption{\textbf{Compute Utilization} simulated across a range of bandwidth. A compute utilization of 0.8 means 80\% of the time is spent in computation, and 20\% in communication. Our best method reaches a compute utilization of 95\% for models 1B, 10B, and 100B with a bandwidth roughly constant between 1 and 5 Gbit/s. Data-Parallel on the other hand requires 100, 200, and 300Gbit/s.}
\label{fig:bandwdith}
\end{figure*}

To highlight the impact of our contributions in a controlled setting, we built a simulator to estimate the \textbf{compute utilization} of each method: how much time is spent doing computation v.s. communication. The simulation is a DAG with four different types of nodes as seen in \autoref{fig:schedule}: forward in blue, backward w.r.t activations and parameters in green, (outer or not, for resp. DiLoCo and Data-Parallel) gradients reduction in purple. Each node represent a single layer. Therefore, the total number of nodes, for a single step, is $4 \times L - 1$ (because we don't need the backward w.r.t activations of the first layer). The overall training is represented by a DAG made of such nodes. We use this simulator to estimate the \textbf{compute utilization} of each method: how much time is spent doing computation v.s. communication. Ideally this number is close to 1.0: No time is spent waiting for communication. It is more useful than just reporting the reduction in data transferred because our overlapping method (\autoref{sec:model_overlapping}) reduces the latency while keeping the amount of data exchanged constant.

In \autoref{fig:schedule}, we report:
\begin{itemize}
    \item \textbf{Data-Parallel}: the baseline which communicates gradients of the full model at every step;
    \item \textbf{DiLoCo}: which communicating outer gradients of the full model  once in a while (in this example, every $H=2$ steps);
    \item \textbf{Streaming DiLoCo}: which communicates outer gradients only for a subset of the model (here the fragment size is a single layer and there are two fragments) every $H=2$ steps;
    \item \textbf{Streaming DiLoCo with overlapping communication and computation}: This is similar to the above but gradients sent across workers are only needed after $\tau$ steps (in this example $\tau=1$).
\end{itemize}

The simulated compute utilization (CU) depends on some factors, listed as columns in \autoref{tab:simulation_hps}. For the model scales 10B and 100B, we estimate step time (pure compute) based on the flops profile, a reasonable MFU ($40\%$), and hardware theoretical flops per seconds. We simulate  training of each method, across three scales (1B, 10B, and 100B) under various bandwidth profiles \autoref{fig:bandwdith}.

We make several observations: 1) Streaming DiLoCo (in \textcolor{ForestGreen}{green}) improves the CU of DiLoCo (in \textcolor{orange}{orange}) despite exchanging as much data, because it reduces the latency by splitting the communication of the outer gradients across fragments. 2) only overlapping communication with computation can reach full $100\%$ compute utilization. 3) the required bandwidth can become {\em lower} as the model scale gets {\em larger} when overlapping communication with computation, because longer compute step time (forward \& backward) will provide more time to perform the synchronization across workers. 

The last point may seem counter-intuitive at first glance, but is the main advantage of our method, exploiting the \textit{square-cube law of distributed training} \citep{ryabinin2023swarmparallelismtraininglarge} where computation scales worse than communication ($O(n^3)$ vs $O(n^2$) for a square matrix $n\times n$). We provide in the appendix, in \autoref{fig:bandwdith_sec}, the simulated compute utilization for a 100B model across various compute step time.

\begin{remark}
Of course this is only a simulation of what we expect to happen in practice. Such simulation is not perfect because for instance we consider only the bandwidth between datacenters and not the local bandwidth between devices. We believe however, that this is still a useful tool\footnote{\url{https://en.wikipedia.org/wiki/Bonini\%27s_paradox}} to  estimate device utilization.
\end{remark}
\vspace{-0.8cm}

\begin{figure*}[t]
\centering
\captionsetup[subfigure]{justification=centering}
\begin{subfigure}{0.45\linewidth}
  \centering
  \includegraphics[width=1\linewidth]{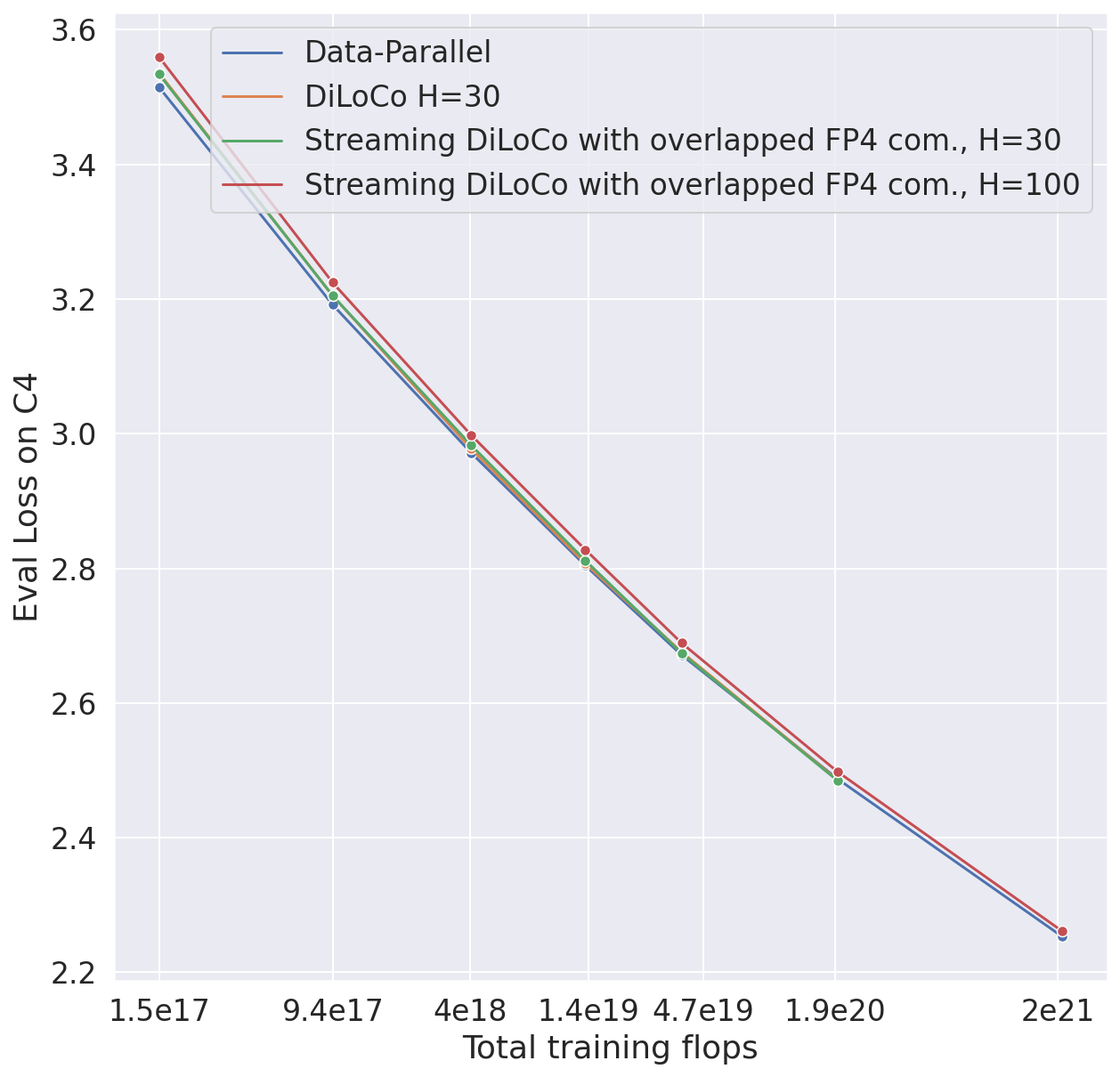}
  \caption{Evaluation loss on C4}
  \label{fig:scaling_loss}
\end{subfigure}\hfill
\begin{subfigure}{0.45\linewidth}
  \centering
  \includegraphics[width=1\linewidth]{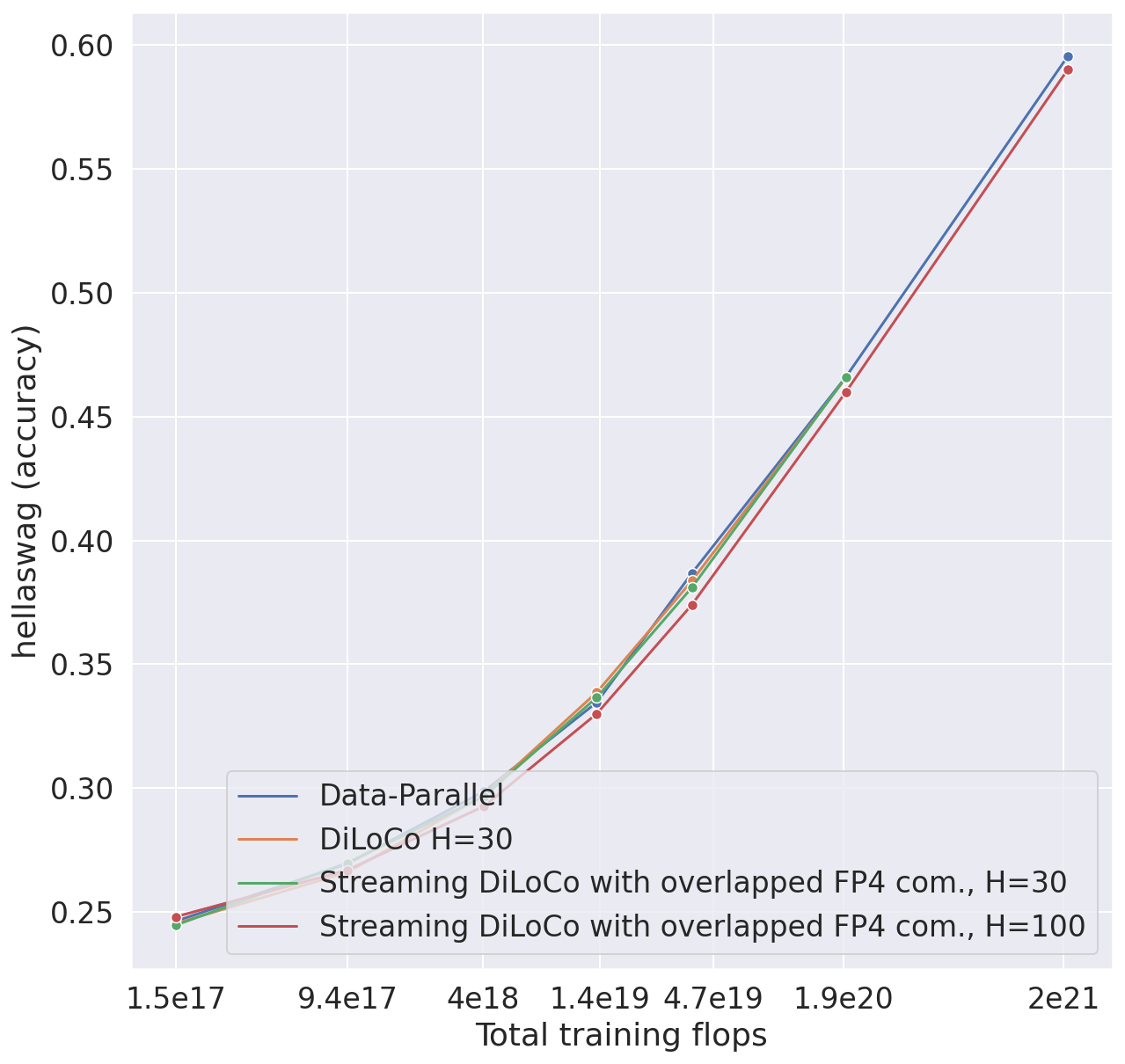}
  \caption{HellaSwag accuracy}
  \label{fig:scaling_hellaswag}
\end{subfigure}
\caption{\textbf{Scaling} models from 35M (1.49e17 flops) to 4B parameters (2e21 flops) on C4.}
\label{fig:scaling}
\end{figure*}

\subsection{LLM Scaling Experiments}
\label{sec:llm_exps}
We perform our experiments with a Chinchilla architecture \citep{hoffmann2022chinchilla}. Following \cite{wortsman2023smallscaleproxieslargescaletransformer} and \cite{jaghouar2024intellect1}, we use QKNorm \citep{henry2020querykeynormalization} and a Z-loss \citep{chowdhery2023palm} with a factor of $1e-4$ to stabilize training. We report in \autoref{tab:hp_architecture} the architecture hyperparameters and token budget at each scale. Unlike recommended in Post-Local SGD \citep{Lin2020_localsgd}, we train all our models from scratch. The main hyperparameter of DiLoCo is its outer learning rate; we tuned it to be optimal at small scale at $0.4$, and kept it fixed across all scales. Likewise, for the simplicity, and to show that Streaming DiLoCo is a drop-in replacement of DiLoCo, we used the same outer learning rate, without further hyperparameters tuning.

Except mentioned otherwise, we use the C4 dataset \citep{c4} and train models from 35 million to 4 billion parameters, all with a sequence length of $1{,}024$. Each scale is trained with the chinchilla-optimal number of steps. We use 2 DiLoCo replicas, each of them performing FSDP \citep{zhao2023fsdp} across their respective closely located devices.

%\mr{I'd cut the rest of this paragraph.}An example setting would be to have one replica in the USA and one in Europe. Note that it is also possible to perform multi-levels of DiLoCo bilevel-optimization, where the DiLoCo replicas themselves have multiple sub-DiLoCo replicas as in Photon \citep{sani2024photonfederatedllmpretraining}. In practice, we don't do hierarchical DiLoCo, but each DiLoCo replica uses locally FSDP \citep{zhao2023fsdp}.

For training we use a modified version of the NanoDO codebase \citep{nanodo} that uses DrJax \citep{rush2024drjax} to parallelize inner steps across replicas. The inner optimization is done with an annotated variant of \texttt{jax.vmap} for the optimization step, with parameters having an extra leading axis for the DiLoCo replicas. The outer optimization is implemented with an all-reduce, without any central parameter server.

\begin{table*}[t]
\centering
\resizebox{1.0\linewidth}{!}{%
%\vspace*{-0.3cm}
\begin{tabular}{@{}l|cccc|cccc@{}}
\toprule
Method & Token Budget & Hours spent w/ $+\infty$ Gbits/s & Hours spent w/ 1 Gbits/s & Terabytes exchanged & Eval Loss $\downarrow$ & HellaSwag $\uparrow$ & Piqa $\uparrow$ & Arc Easy $\uparrow$ \\
\midrule
\multirow{3}{*}{Data-Parallel} & 25B & 0.67 & 109 & 441 & 2.67 & \textbf{42.09} & 67.35 & \textbf{40.42} \\
 & 100B & 2.7 & 438 & 1,767 & 2.52 & 49.78 & 69.15 & \textbf{44.03} \\
 & 250B & 6.75 & 1097 & 4,418 & \textbf{2.45} & 53.86 & 70.45 & \textbf{44.21} \\
 \midrule
\multirow{3}{*}{\parbox{3.5cm}{\centering Streaming DiLoCo \\ with overlapped FP4 communication}} & 25B & 0.67 & 0.88 & 1.10 & \textbf{2.66} & 42.08 & \textbf{67.46} & 38.42 \\ 
& 100B & 2.7 & 3.5 & 4.42 & \textbf{2.51} & \textbf{49.98} & \textbf{69.96} & \textbf{44.03} \\
& 250B & 6.75 & 8.75 & 11.05 & \textbf{2.45} & \textbf{54.24} & \textbf{71.38} & 41.92 \\

\bottomrule
\end{tabular}
}
\caption{\textbf{Overtraining} Data-Parallel and our method on Dolma with a 1 billion parameters model. The latter performs slightly better despite exchanging in total $400\times$ fewer bits, reducing the peak bandwidth by $8\times$, and with a significantly relaxed training communication latency constraint: allow communication to be as long as a full compute step.}
\label{tab:overtrain_steps}
\end{table*}

\subsubsection{Scaling} \label{sec:scaling}

We perform scaling experiments on C4, with models ranging from 35 millions parameters to 1 billion parameters, all with a sequence length of $1{,}024$. For Data-Parallel and Streaming DiLoCo with $H=100$, we also provide results on a 4 billion parameter model. At each scale, we use the Chinchilla-optimal \citep{hoffmann2022chinchilla}  number of steps. We highlight in \autoref{fig:scaling} the evaluation loss (lower is better) and HellaSwag \citep{zellers2019hellaswagmachinereallyfinish} accuracy (higher is better).

First, we observe in \autoref{fig:scaling} that Data-Parallel (in \textcolor{blue}{blue}), DiLoCo with $H=30$ inner steps (in \textcolor{orange}{orange}), and Streaming DiLoCo with $H=30$ (\textcolor{ForestGreen}{green}) perform all similarly across both loss (at 1B parameters, respectively 2.49, 2.49, and 2.48) and accuracy metrics (resp. 46.6\%, 46.5\%, and 46.6\%). Streaming DiLoCo with more inner steps $H=100$ (in \textcolor{BrickRed}{red}) has slightly worse performance initially but use significantly less bandwidth and the loss improves proportionally better as we scale: scaling law slope for Data-Parallel is -0.13149 while -0.13539 for Streaming DiLoCo. We report in the appendix \autoref{tab:scaling} all metrics, and include two more downstream tasks: Piqa \citep{bisk2019piqareasoningphysicalcommonsense} and Arc-Easy \citep{clark2018arc}. Moreover \autoref{tab:scaling_m4} considers an increased number of DiLoCo replicas.

\subsubsection{Overtraining on Dolma} \label{sec:overtraining}

The previous experiments were performed on C4 dataset using the chinchilla-optimal number of tokens. Using a 1 billion parameter model, this yields a token budget of 25 billion. However, language models are now usually  \textit{overtrained} \citep{gadre2024languagemodelsscalereliably}. Therefore we perform a comparison of a Data-Parallel baseline vs our full model (streaming DiLoCo with overlapped FP4 communication) on the Dolma dataset \citep{soldaini2024dolmao} with a 1 billion parameter model and with a token budget of 25, 100, and 250 billions tokens (resp. 1.9e20, 7.6e20, and 1.9e21 flops) using a sequence length of $2{,}048$. In that larger, more realistic setting, we set the number of inner steps between synchronization  to $H=100$ to further minimize communication. 

We report results in \autoref{tab:overtrain_steps}, and note that both our method and the baseline perform similarly w.r.t loss and accuracy on downstream tasks (HellaSwag, Piqa, Arc-Easy). In addition of being neutral in term of ML performance: 1) the amount of bits exchanged between non-colocated devices over the course of training is $400\times$ higher for Data-Parallel; 2) the peak bandwidth (amount of bits exchanged at given moment) is reduced by $\frac{\text{num layers = } 24}{\text{fragment size = } 3} = 8\times$; and 3) while Data-Parallel ideally hopes for a 0 second latency when communicating, our overlapping scheme allows us a latency as long as a full forward/backward pass, which is several seconds at large scale. For those reasons, we believe our work is step towards a truly ``\textit{distributed free lunch}''.

\subsection{Ablations} \label{sec:ablations}

To ablate the importance of each component of Streaming DiLoCo, we perform all our ablations on  a model of size 500 million parameters using the C4 dataset with the chinchilla-optimal number of steps and a token budget of 11 billions.

We split our ablations section in three parts, corresponding to the three improvements brought in this paper: namely 1) Streaming synchronization in section \ref{sec:ablation_streaming}, 2) overlapping communication with computation in section \ref{sec:ablation_overlap}, and 3) finally quantized communication in section \ref{sec:ablation_quant}.

\subsubsection{Ablating the streaming synchronization}\label{sec:ablation_streaming}

In this ablation section, we consider different settings for streaming DiLoCo presented in \autoref{sec:model_streaming}.

\begin{figure}[t]
\captionsetup[subfigure]{justification=centering}
\begin{subfigure}{1.0\linewidth}
  \centering
  \includegraphics[width=1\linewidth]{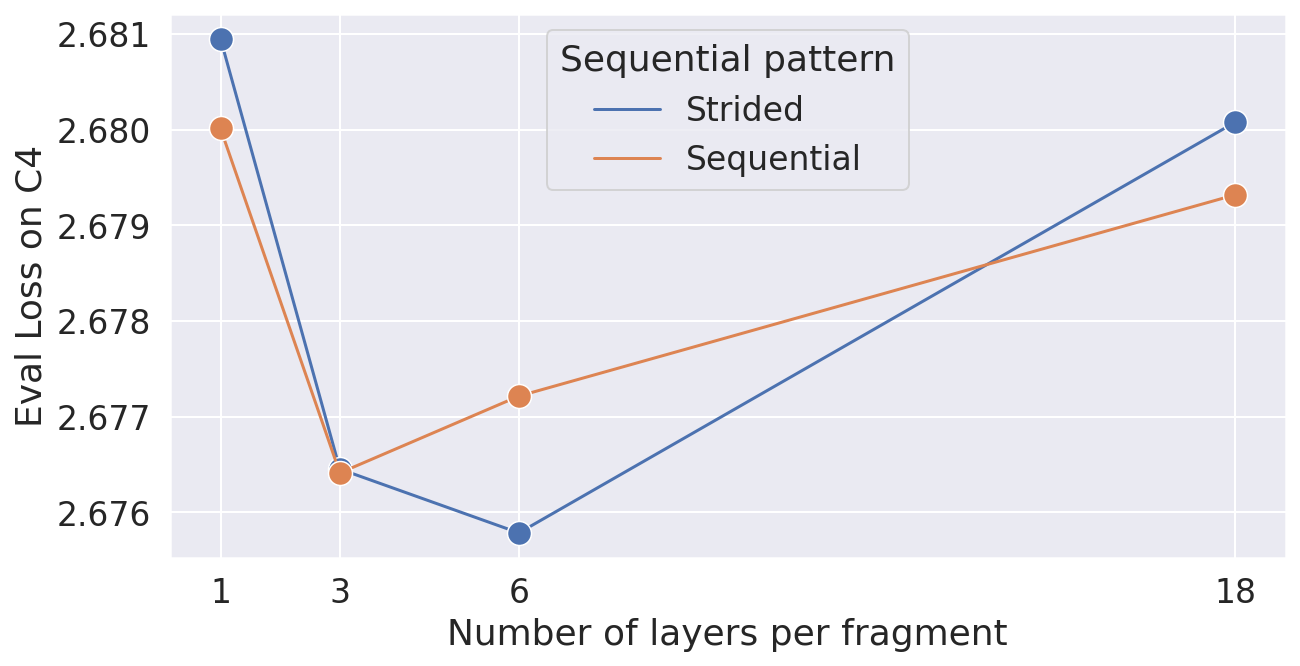}
  \caption{C4 eval loss}
  \label{fig:fragment_size_loss}
\end{subfigure}
\\
\begin{subfigure}{1.0\linewidth}
  \centering
  \includegraphics[width=1\linewidth]{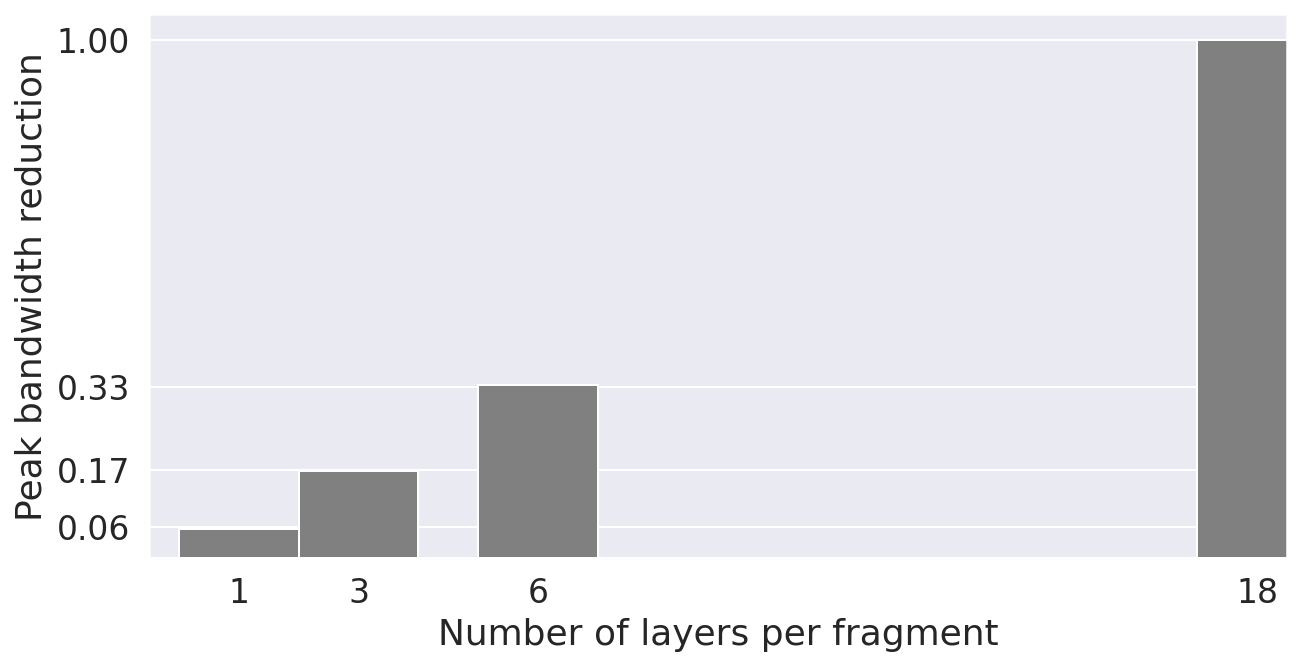}
  \caption{Peak bandwidth reduction}
  \label{fig:fragment_size_bandwidth}
\end{subfigure}
\caption{\textbf{The fragment's size} will determine the peak bandwidth but also the learning dynamics. We choose in practice 3 layers per fragment across all model scales.}
\label{fig:fragment_size}
\end{figure}

\begin{figure}[t]
  \centering
  \includegraphics[width=1\linewidth]{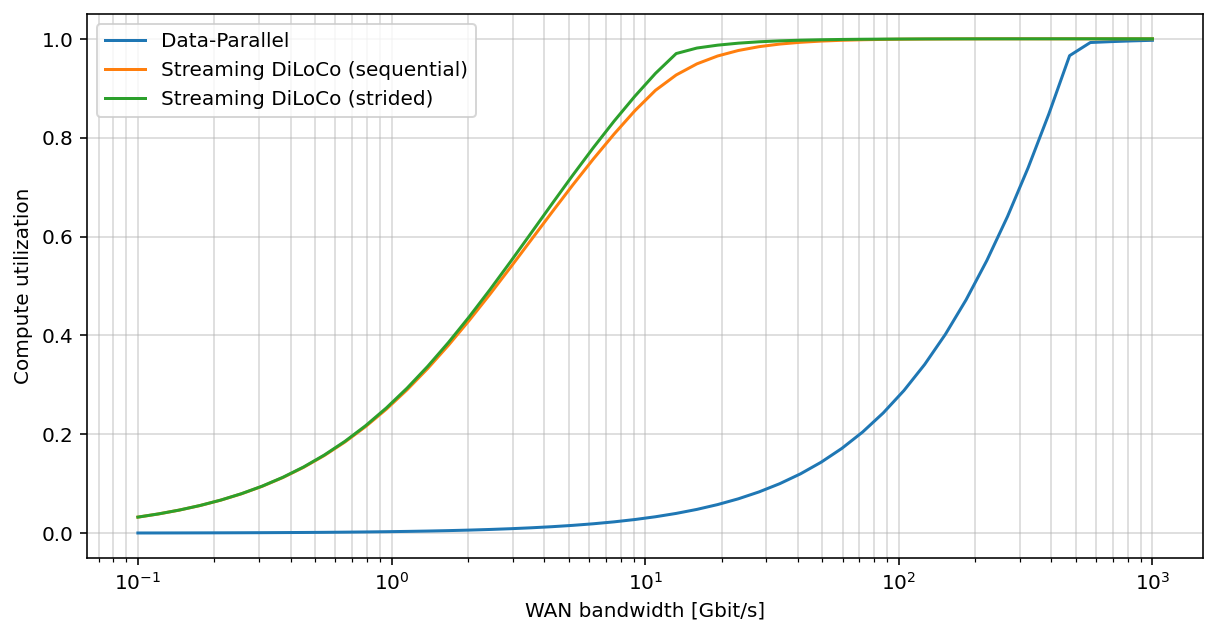}
\caption{\textbf{Compute utilization profile} of sequential vs strided pattern for a 100 billion parameters model.}
\label{fig:bandwidth_100b_stride}
\end{figure}

\paragraph{Number of synced layers per fragment.} We ablate in \autoref{fig:fragment_size} the fragment size, \textit{i.e.}, how many transformer blocks are included in a fragment. Based on this analysis, we choose a fixed fragment size of 3 layers, striking a desirable trade-off between ML performance and reduction of peak bandwidth (for which the smaller the fragments the better). We also consider whether to have a \texttt{sequential} or \texttt{strided} pattern (see illustration in \autoref{fig:streaming_pattern} for a reference). We choose the latter for several reasons: 1) ML performance is slightly better for the fragment size we consider, 2) deeper networks, with a small fragment size (e.g. 3), should benefit more from striding by spreading out up-to-date synchronized layers across the full depth of the network. Finally, 3) it slightly improves the compute utilization (see \autoref{fig:bandwidth_100b_stride}) by allowing better overlapping schedule, as clearly seen in \autoref{fig:schedule_strided} in the appendix.

\paragraph{Comparison to FedPart.} Streaming DiLoCo bears similarity with  concurrent work dubbed FedPart \citep{wang2024fedpart}, where a subset of the model is also exchanged at each round. However, FedPart argues that non-shared layers should be be frozen during inner optimization. We believe this is rather flops-inefficient: For an 18 layer model, with 3 layers per fragment, 15 layers (83\%) are frozen at any point in time despite doing forward/backward computation. We ran comparison of Streaming DiLoCo with and without the frozen pattern proposed by FedPart, reaching respectively on the C4 eval loss $3.2145$ and $2.6749$. Freezing the $18-3=15$ layers that won't be synchronized at the given round therefore results in a 20\% increase of the evaluation loss. These results confirm our intuition that while freezing layers may help merging, this incurs a significant flop inefficiency, which might not be acceptable in training-compute bound settings (which are typical in current large scale training of LLMs).

\subsubsection{Ablating the communication overlap}\label{sec:ablation_overlap}

In this ablation section, we  investigate how to overlap  communication with computation, see \autoref{sec:model_overlapping} for reference.

\begin{figure}[t]
  \centering
  \includegraphics[width=1\linewidth]{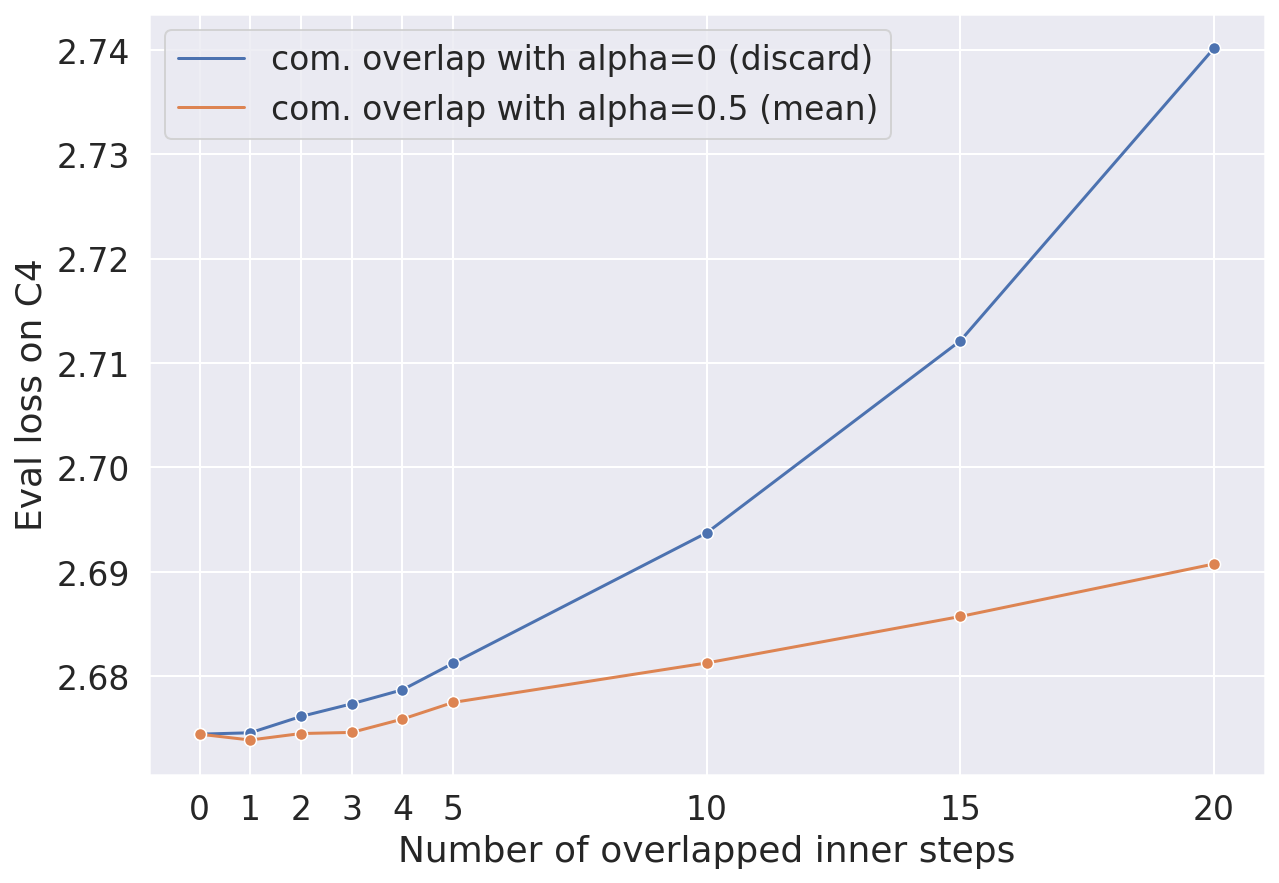}
\caption{Varying \textbf{the number of overlapped inner steps $\tau$} for $\alpha=\{0, 0.5\}$. A larger $\tau$ requires a significantly lower bandwidth, see also \autoref{fig:bandwidth_100b_overlap}.}
\label{fig:num_overlap}
\end{figure}

\begin{figure}[t]
  \centering
  \includegraphics[width=1\linewidth]{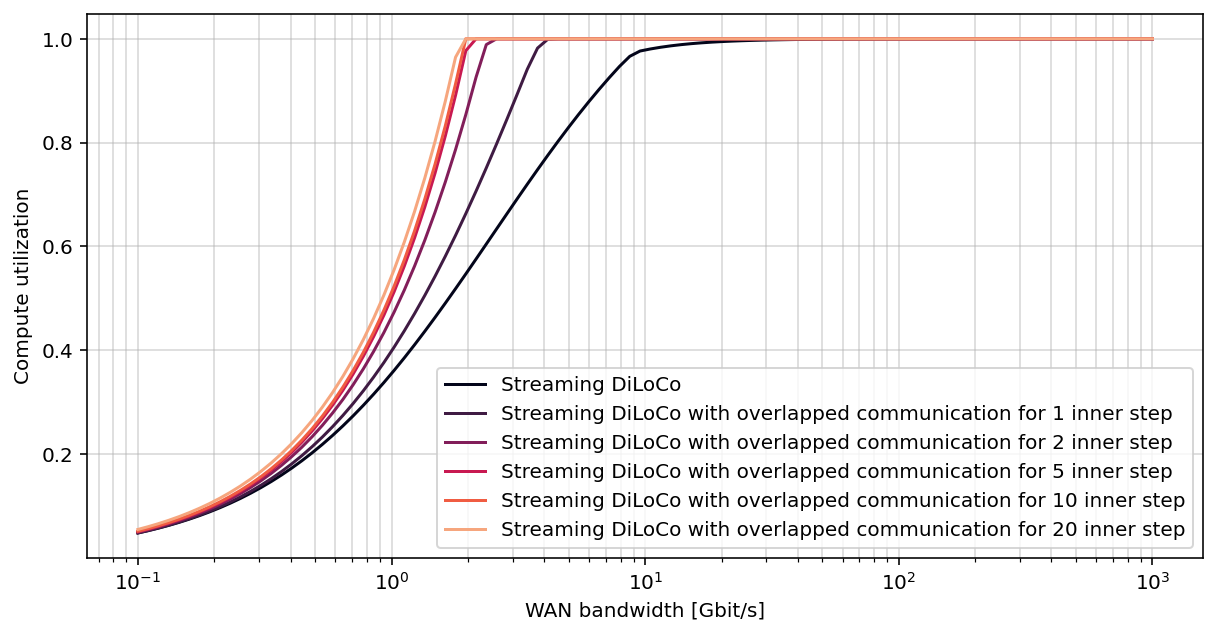}
\caption{\textbf{Estimated compute utilization} for a 100B model  when increasing $\tau$, the number of inner steps which overlap with communication.}
\label{fig:bandwidth_100b_overlap}
\end{figure}

\paragraph{Overlapping.} We first vary the number of inner steps, $\tau$ we use to overlap communication with computation (see \autoref{sec:model_overlapping}). \autoref{fig:num_overlap} shows results varying $\tau$ from 1 to 20, with $\alpha=0$ (discarding any intermediary inner updates) and $\alpha=0.5$ (uniform merging). We can see that the degradation in  evaluation loss is negligible up to an overlap of 10 inner steps ($< 0.2\%$). 
By checking the compute utilization of a 100B model in \autoref{fig:bandwidth_100b_overlap}, we observe little gain in compute time after an overlap of 5 inner steps. Therefore, we advise practitioners to limit the overlap to 5 inner steps. In our main experiments, we used 1 step for simplicity. 

\begin{figure}[t]
  \centering
  \includegraphics[width=1\linewidth]{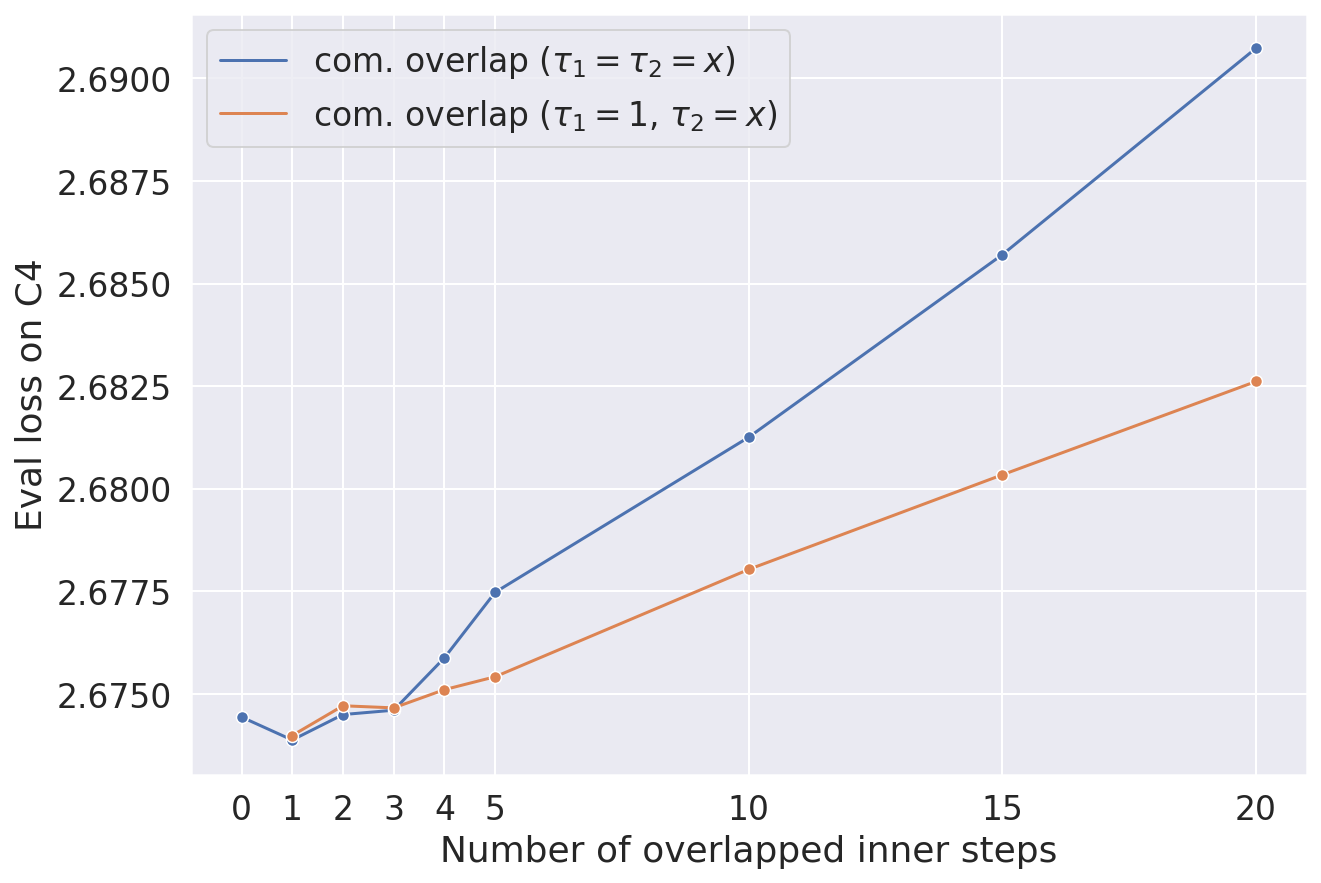}
\caption{Varying \textbf{the number of overlapped inner steps $\tau_2$} for the second worker while keeping $\tau_1=1$. For all data points, $\alpha=0.5$. Training is very robust for values of $\tau_2$ less than 5.}
\label{fig:num_overlap_async_between_workers}
\end{figure}

\paragraph{Overlapping with some slack between workers.} 
If workers use heterogeneous device types (e.g. TPUv5e vs TPUv6e) or are just placed in different environments, their execution speed might vary and it would be inefficient to force them to synchronize at the same optimization step.
In this case, we could grant workers with some slack. For instance, in a 2 DiLoCo replicas setting, both workers could send their respective outer gradients at the same step (but not necessarily at the same time) and they could receive the update at a different step \citep{liu2024asyncdiloco}. We accomplish this by simply using a different $\tau$ per worker ($\tau_1$ and $\tau_2$) as shown in line 7-9 in \autoref{alg:streaming}. We show in \autoref{fig:num_overlap_async_between_workers} the evaluation loss when $\tau_1 = 1$ and varying $\tau_2$. Similarly to \autoref{fig:num_overlap}, the loss degradation is limited under a delay of up to 5 inner steps. This result suggests that Streaming DiLoCo is rather robust and could support training of large models on several heterogeneous workers.

\subsubsection{Ablating the quantized communication}\label{sec:ablation_quant}

Finally, in this section, we consider various schemes to quantize our communication, as proposed in \autoref{sec:model_low_precision}.

\begin{figure}[t]
\captionsetup[subfigure]{justification=centering}
\begin{subfigure}{0.95\linewidth}
  \centering
  \includegraphics[width=1\linewidth]{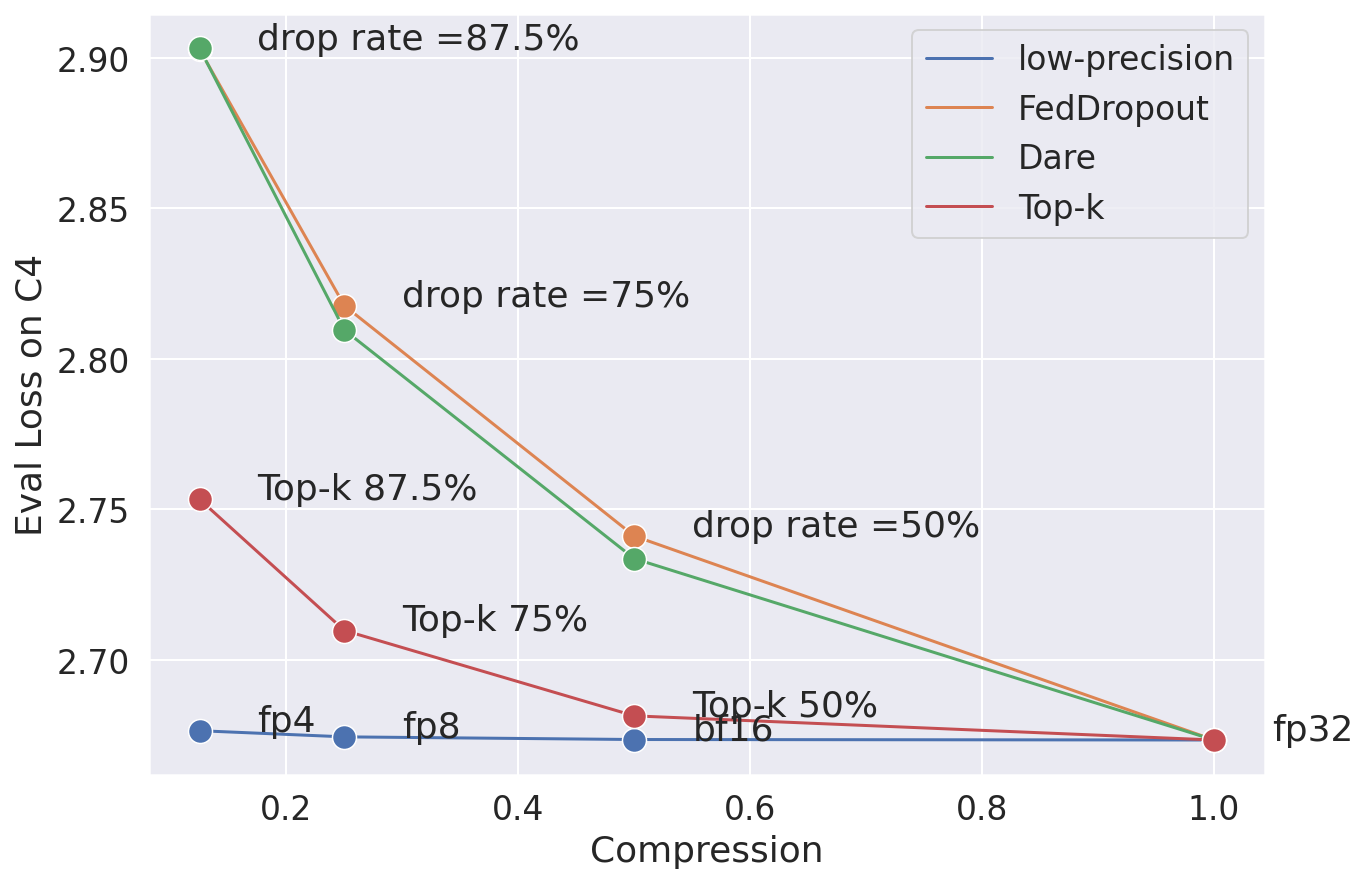}
  \caption{C4 evaluation loss}
  \label{fig:compression_loss}
\end{subfigure}
\\
\begin{subfigure}{0.95\linewidth}
  \centering
  \includegraphics[width=1\linewidth]{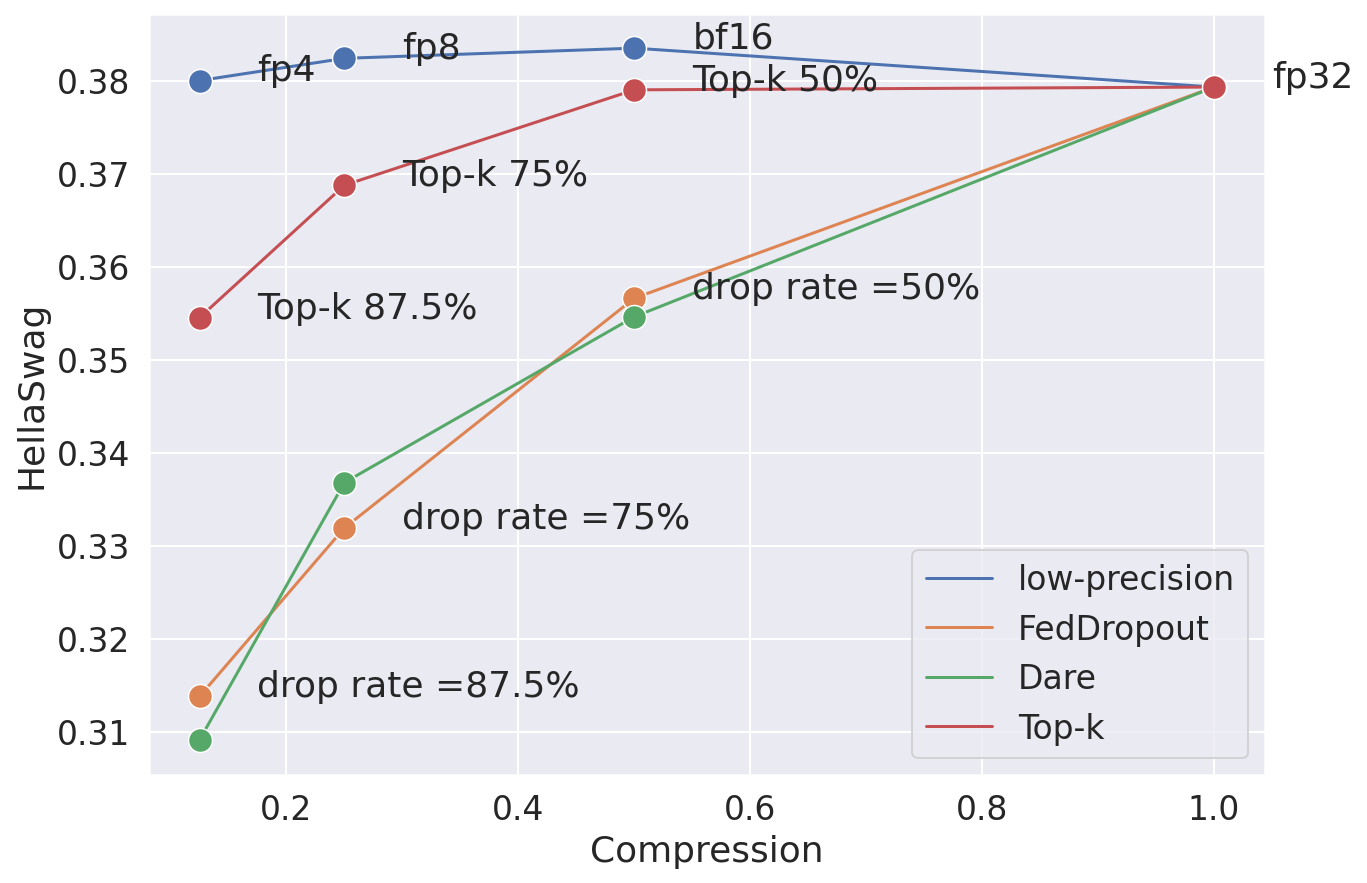}
  \caption{HellaSwag accuracy}
  \label{fig:compression_hellaswag}
\end{subfigure}
\caption{\textbf{Compressing the outer gradients} with either value dropping (FedDropout, Dare) or using lower-precision floating point numbers.}
\label{fig:compression}
\end{figure}

\paragraph{Compressing outer gradients.} We ablate in \autoref{fig:compression} two ways of compressing the outer gradients: either by setting to zero some values (FedDropout \citep{wen2022feddropout}, Dare \citep{yu2024dare}, and Top-K selection) or by lowering the precision.  In all cases, we accumulate the outer update in float32.  We report both the evaluation loss on C4 and the accuracy on HellaSwag. Interestingly, lowering the precision, from float32 to float4 does not affect the performance, while setting some values to zero is significantly worse, particularly when zero-ing out at random. We also considered Ties-Merging's pruning method \citep{yadav2023tiesmerging} but preliminary experiments showed it also underperformed; however this approach might become advantageous with larger number of replicas $M$.

\section{Related Works} \label{sec:related}

\paragraph{Model merging.} Model merging, a subfield within the broader study of linear mode connectivity, explores the potential of linearly interpolating between parameters of multiple models to synthesize a unified model that inherits the strengths of its constituents \citep{matena2022fisherweightedaveraging,wortsman2021learning}.  A key finding in this domain is the existence of low-loss pathways within the parameter space \citep{frankle2020linear,neyshabur2021transferredtransferlearning} that connect independently trained models, effectively circumventing the anticipated loss barriers. For instance, \citet{wortsman2022soup} demonstrated that averaging the parameters of models fine-tuned from a common pre-trained initialization on diverse tasks \citep{rame2023modelratatouillerecyclingdiverse} or with varying hyperparameters \citep{wortsman2022soup} yields a performant merged model. This approach, initially demonstrated in computer vision, has been successfully extended to natural language processing \citep{li2022branchtrainmerge}, reinforcement learning with human feedback \citep{rame2023rewarded}, noisy label learning \citep{rebuffi2022revisiting}, and out-of-distribution generalization \citep{rame2023diverse}. Recent research has further investigated alternative strategies for mitigating loss barriers in model merging, including techniques based on parameter space transformations and other model surgery methods aiming to resolve merging conflicts \citep{jordan2023repair,stoica2023zipit,jin2023dataless,ainsworth2023gitrebasinmergingmodels,yadav2023tiesmerging,yu2024dare}.

\paragraph{Federated learning / local SGD.} While model merging proposes to combine several models once, FedAvg \citep{mcmahan2017fedavg} and Local SGD \citep{stich2019local} do it multiple times with the goal of reducing bandwidth requirements: they operate by performing local training (typically via SGD) across workers for some number of steps before doing some kind of synchronization of worker parameters, or aggregation of parameters across workers. In their original forms, both FedAvg and Local SGD simply averaged the parameters across workers. As shown by \citet{reddi2021adaptive}, the synchronization is more effective when each worker calculates a ``model delta'', and these are aggregated over workers to produce a pseudo-gradient ~\citep{reddi2021adaptive,ilharco2022patching} or \textit{outer gradient}, which is then fed to a first-order optimizer. This yields a bi-level optimization scheme with inner optimizers and an outer optimizer, referred to as FedOpt by \citet{reddi2021adaptive}, who propose using SGD as the inner optimizer and adaptive methods like Adam~\citep{kingma2014adam} as the outer optimizer in resource-constrained FL settings.

\paragraph{Distributed training for LLMs.} The increased requirements of training large language models (LLMs) hastened the need for distributed methods, for both inference \citep{borzunov2023petals} and training \citep{presser2020stub,diskin2021distributedcollab,ryabinin2021moshpit}. More recently, DiLoCo \citep{douillard2023diloco} proposed a particular instantiation of FedOpt \citep{reddi2021adaptive} with AdamW \citep{loshchilov2018adamw} as inner optimizer and Nesterov \citep{sutskever2013nesterov} as outer optimizer \citep{huo2020outernesterov}. This simple formulation proved to be effective for distributed training with LLMs, where the number of replicas is small (<100) and without replica sampling, closer to cross-silo federated learning \citep{kairouz2021advances}. The FedOpt algorithm was also shown to be effective in training LLMs in settings that looked more like cross-device federated learning~\citep{charles2024towards}. The empirical success of DiLoCo has been reproduced multiple times \citep{jaghouar2024opendiloco,sani2024futurelargelanguagemodel} and has been successfully scaled up to 10 billion parameter models \citep{jaghouar2024intellect1}. Related, a simple change on how the outer Nesterov accumulates outer gradients proved to handle well asynchronicity between workers of different speeds \citep{liu2024asyncdiloco}. DiLoCo adds a new axis of paralellism to distributed training \citep{shoeybi2020megatronlmtrainingmultibillionparameter}, and is compatible \citep{jaghouar2024intellect1} with other existing axes like FSDP \citep{zhao2023fsdp}, or even another level of federated learning \citep{sani2024photonfederatedllmpretraining}.

% \cite{ortiz2021tradeoffs} is one of the few works in federated learning / local SGD body of literature that has validated on a large-scale setting. They consider ImageNet \citep{deng2009imagenet} with Resnet50 and Resnet101 \citep{he2015resnet}, and found that local SGD struggles at scale. In particular, they reported that fewer inner steps (e.g., $H=8$), no pretraining, and a relatively large number of replicas ($\ge k=16$) degrade  generalization. Thus the authors conclude that "\textit{local SGD encounters challenges at scale.}". Instead, we show in \autoref{sec:experiments} that \Model{} can robustly operate while communicating $125\times$ less ($H=1000$), even without pretraining, and using up to $4\times$ more replicas ($k=64$) both in the i.i.d. and non-i.i.d. settings. Recently, multiple works \citep{presser2020stub,diskin2021distributedcollab,ryabinin2021moshpit} also applied Local SGD for language models but without outer optimization.

\paragraph{Partial communication.} Communicating a subset of the network is often used in federated learning to provide \textit{personalized} models per user, see FedPart \citep{arivazhagan2019federatedlearningpersonalizationlayers}. DiPaCo \citep{douillard2024dipaco} recently proposed a distributed mixture-of-experts where subsets of the model is synchronized with subsets of the replicas, according to a sharing pattern that is optimized with Expectation-Maximization style of algorithm during training. WASH \citep{fournier2024washtrainensemblecommunicationefficient} and later Sparta \citep{exo2025sparta} propose to frequently exchange a random subset of the neurons. Finally FedPart \citep{wang2024fedpart}, developed at the same time as Streaming DiLoCo, also proposes to share per-layer fragments. However, they argue that for a given communication round, non-shared fragments should not undergo inner optimization, a strategy which we show slows down convergence. Note that all partial communication methods can be seen as a form of \textit{structured} (outer) gradients compression.

\paragraph{Gradient compression.} Data-Parallel (with gradients) and Federated learning (with outer gradients) often share similar methods to compress the communication \citep{lin2020deepgradientcompressionreducing}: from randomly dropping values \citep{wen2022feddropout}, to combining multiple compression schemes (e.g. dropping, top-k, low-precision) \citep{wang2023cocktailsgd}, to use low-rank compression \citep{vogels2020powersgd,zhao2024galore}, or recently to keep only the fast moving components with DCT but communicates via an all-gather collective instead of an all-reduce \citep{peng2024demod}.

\section{Conclusion and Future Work} \label{sec:conclusions}

In this paper, we introduced three improvements over DiLoCo: we synchronize a only subset of the parameters at a time, we overlap the communication of this synchronization over several computation steps, and we compress the outer gradients to communicate to low-precision with only four bits. All these innovations combined together leads to a training with similar ML-performance as a classical Data-Parallel training, while using $400\times$ less bandwidth, reducing the peak bandwidth compared to DiLoCo's bursts of communication, and allowing communication to have an ideal non-zero latency by overlapping it with computation.

In sum, we can reach a similar compute utilization as the widely used Data-Parallel using two orders of magnitude less Gbit/s bandwidth, while performing comparably in term of training loss and downstream evaluation accuracies as Data-Parallel. For those reasons, we claim that this work in a first step towards what we call a \textit{distributed} free lunch, paving the way for a new way to train distributed networks with reduced bandwidth and yet without trading-off model quality.

\paragraph{Next.} In our view, the ubiquity of co-located Data-Parallel training is likely due to the hardware lottery \citep{hooker2020hardwarelottery}, when ``\textit{a research idea wins because it is suited to the available software and hardware and not because the idea is superior to alternative research directions}''. Data-Parallel training has been extensively studied, tuned, and scaled \citep{kaplan2020scalinglawsneurallanguage}, and it is hard to beat the wisdom-of-the-crowd of thousands of researchers.  In contrast, the federated learning literature has mainly studied smaller scale models, primarily due to its focus on edge devices. There are huge opportunities for bringing the ideas from the federated learning literature to the new world of large scale training for LLMs. A critical next work is to study how new distributed methods like ours should be tuned and scaled across multiple axes (e.g. model size, overtraining factor, number of replicas). In particular, how to scale efficiently the number of DiLoCo replicas given an equivalent token budget is most needed. 

More generally, reducing the communication problem to a minor obstacle allows new classes of co-designed architectures and training paradigms (for example  \citet{douillard2024dipaco}) maximizing available compute \citep{sutton2019bitterlesson}: we hope to see the training of modular constellations of small models loosely connected \citep{dean2021pathways} across heterogeneous devices, using compute arbitrage spread world-wide.  
\section*{Acknowledgements}
We would like to thank Alban Rrustemi, Jeff Dean, Michael Isard, Sebastian Borgeaud, Rohan Anil, Koray Kavukcuoglu, and Raia Hadsell for their feedback and leadership support; Andrei Rusu, Adhiguna Kuncoro, Lucio Dery, Rachita Chhaparia, Zohar Yahav, Qixuan Feng, Zack Nado, Nova Fallen, Nicole Mitchell, Sean Augenstein, and Stephen Roller for the helpful advices; finally Alberto Magni, Juliana Vincente Franco, Joel Wee, James Lotte, Matthew Johnson, and Blake Hechtman for unblocking us through so many engineering hurdles alongside our journey.
%\newpage
%\clearpage
% Bibliography components
\raggedbottom
\bibliographystyle{plainnat}
\nobibliography*
\bibliography{main}

\clearpage
\onecolumn
\section*{Supplementary Materials}\label{sec:supp}

\paragraph{Architecture hyperparameters.} We detail the architecture across model scales in \autoref{tab:hp_architecture}. The token budget per scale is computed from the chinchilla-optimal amount of flops \citep{hoffmann2022chinchilla}.

\begin{table}[!ht]
\centering
\resizebox{1.0\linewidth}{!}{%
%\vspace*{-0.3cm}
\begin{tabular}{@{}l|cccc@{}}
\toprule
Model scale & Hidden dim & Num layers & Num heads & Token budget \\
\midrule
35M & $2{,}048$ & 6 & 8 & 700M\\
100M & $3{,}072$ & 9 & 12 & 1.5B \\
200M & $4{,}096$ & 12 & 16 & 3.5B \\
300M & $5{,}120$ & 15 & 20 & 6B\\
500M & $6{,}144$ & 18 & 24 & 11B \\
1B & $8{,}192$ & 24 & 32 & 25B\\
4B & $12{,}288$ & 36 & 48 & 83B\\
\bottomrule
\end{tabular}
}
\caption{\textbf{Architecture hyperparameters}: we consider model from 35M to 4B with the following hyperameters and chinchilla-optimal token budget. For all model scale, the vocabulary size is $32{,}000$.}
\label{tab:hp_architecture}
\end{table}

\begin{figure}[t]
\captionsetup[subfigure]{justification=centering}
\begin{subfigure}{0.95\linewidth}
  \centering
  \includegraphics[width=1\linewidth]{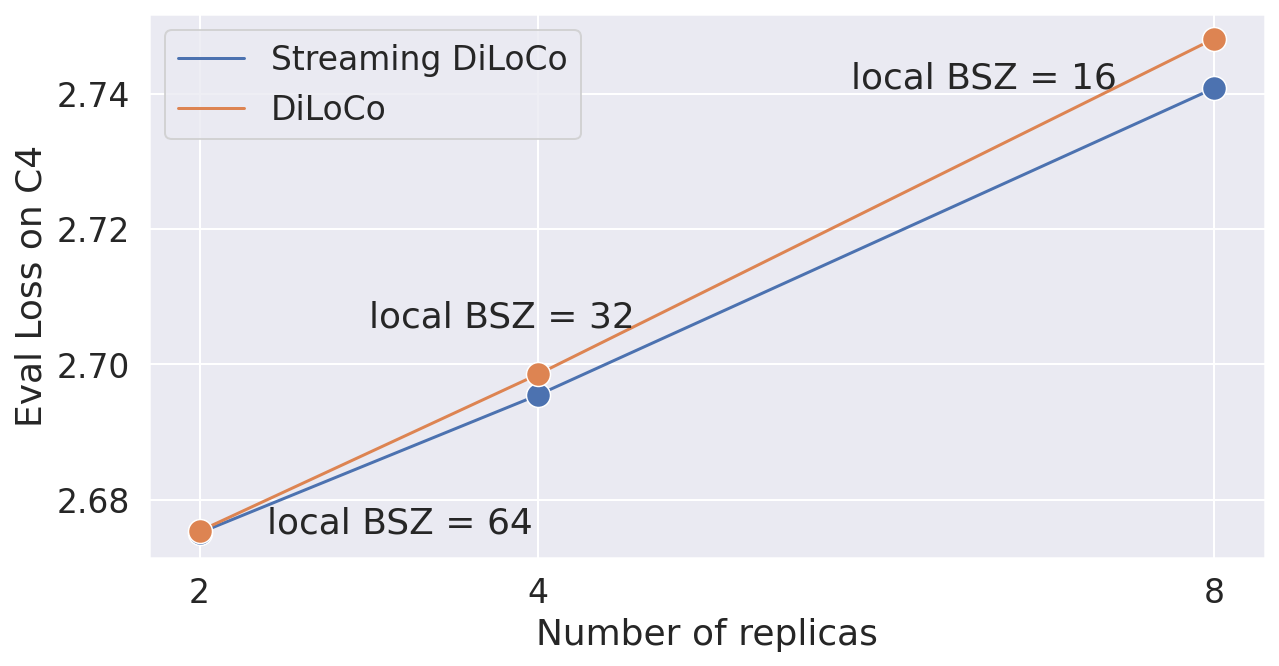}
\caption{Keeping the \textbf{\textit{global} batch size constant}, and thus decreasing the \textit{local} per-replica batch size.}
\label{fig:num_replicas_cst_global}
\end{subfigure}
\\
\begin{subfigure}{0.95\linewidth}
  \centering
  \includegraphics[width=1\linewidth]{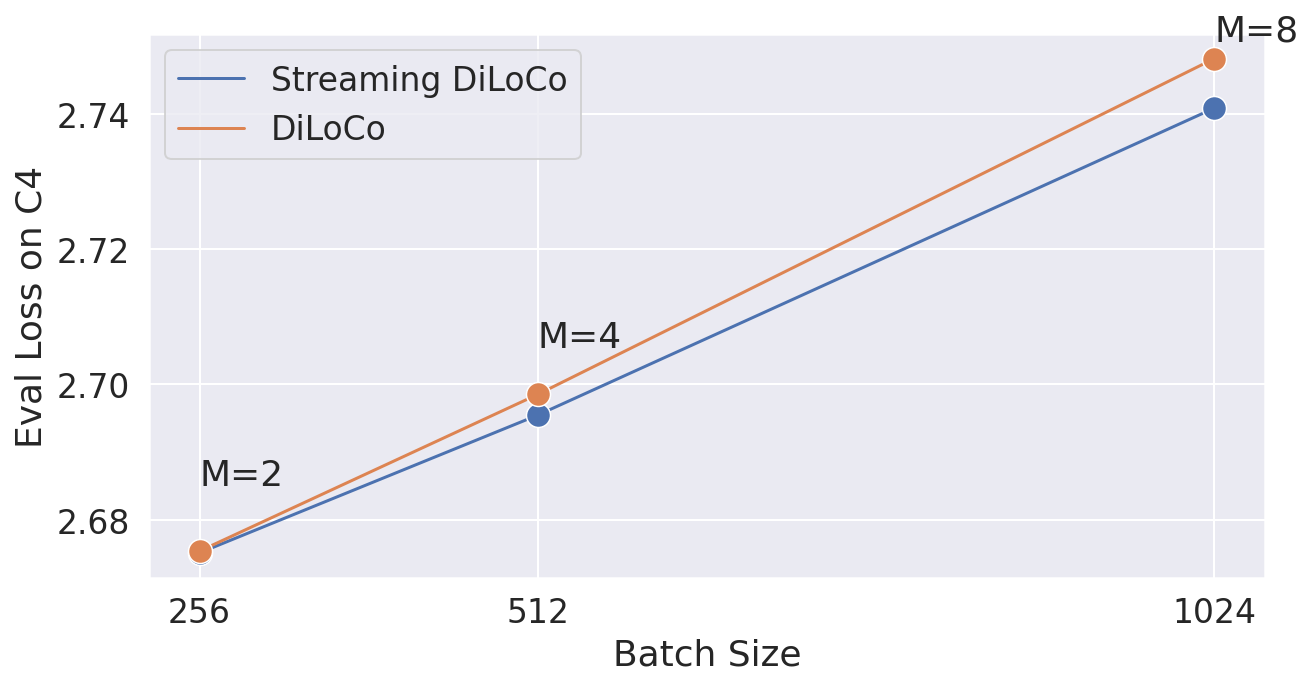}
\caption{Keeping the \textbf{\textit{local} per-replica batch size constant}, and thus increasing the \textit{global} batch size.}
\label{fig:num_replicas_cst_local}
\end{subfigure}
\caption{\textbf{Scaling the number of DiLoCo replicas $M$} from $M=2$ to $M=4$. For all experiments, the token budget is kept constant.}
\label{fig:num_replicas}
\end{figure}

\paragraph{Number of replicas.} We perform our main experiments with 2 replicas for simplicity but showcase in \autoref{fig:num_replicas} an ablation of DiLoCo vs Streaming DiLoCo where the number of replicas $M$ vary from 2 to 8. Contrarely to \citep{douillard2023diloco}, we keep the total token budget constant. In \autoref{fig:num_replicas_cst_global}, we keep the global batch size constant, and thus reduce the local per-replica batch size). In \autoref{fig:num_replicas_cst_local}, we keep the local per-replica batch size constant, and thus increase the global batch size but also reduce the total number of steps.

\begin{figure}[t]
  \centering
  \includegraphics[width=1\linewidth]{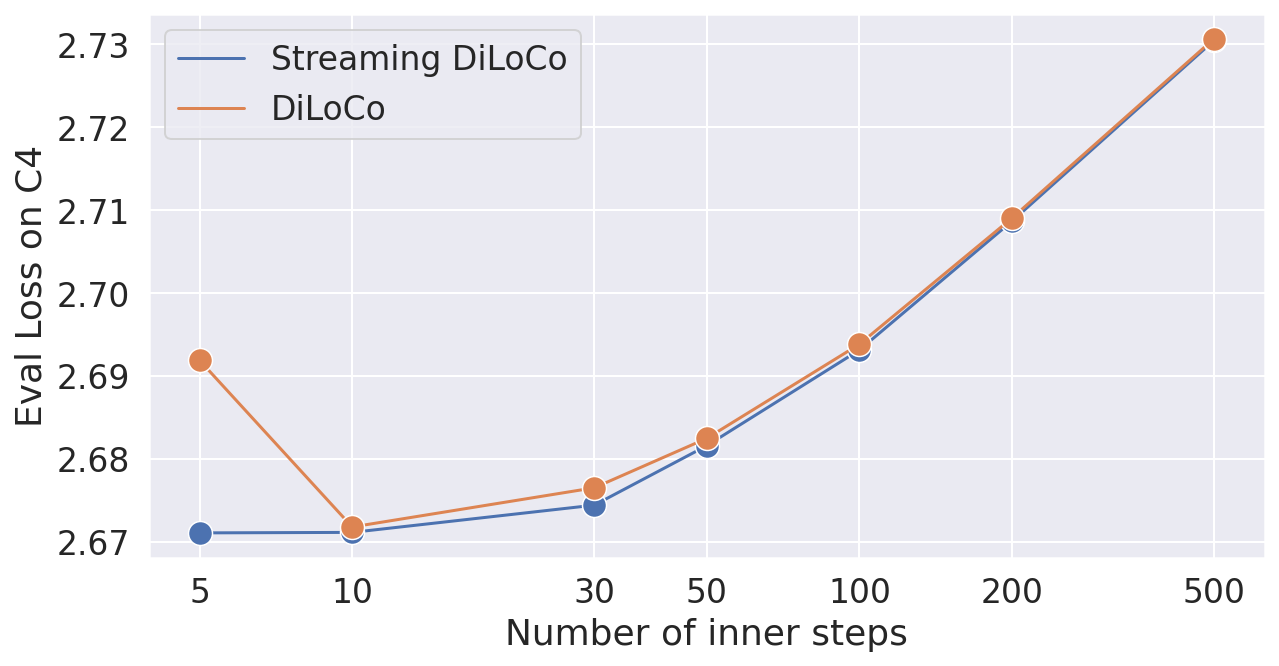}
\caption{Varying \textbf{the number of inner steps $H$} for DiLoCo and Streaming DiLoCo while keeping the total number of steps constants. A lower $H$ means more communication rounds to be done.}
\label{fig:num_inner_steps}
\end{figure}

\paragraph{Number of inner steps.} The number of inner steps $H$, has an engineering effect and a learning effect: a larger $H$ means less frequent synchronization and thus less required bandwidth. On the other hand, a too small $H$ produce noisy small-normed outer gradients and a too high $H$ will see replicas drifting apart. Therefore, some middle ground needs to be found. We ablate in \autoref{fig:num_inner_steps} and find that while Streaming DiLoCo has similar behavior as DiLoCo when $H$ increases, it is more robust to low values of $H$.

\paragraph{Which parameters to evaluate.} We considered multiple subset of the parameters to use for evaluation: 1) the arbitrarily chosen first replica ($\theta_1$), 2) an average of all replicas ($\frac{1}{M} \sum_{m=1}^M \theta_m$), or 3) the globally shared outer parameters ($\theta$). Note that the latter is made of fragments that were synchronized at different points in time. We show the performance of each subset in \autoref{tab:evaluate_mode}: The difference here between these methods is small, but the outer parameters yield slightly better performance. 
%This is also in line of what we saw across the development of our methods.

\begin{table}[t]
\centering
%\resizebox{1.0\linewidth}{!}{%
%\vspace*{-0.3cm}
\begin{tabular}{@{}l|cc@{}}
\toprule
Parameters evaluated & Eval Loss & HellaSwag \\
\midrule
First replica & 2.77 & 37.77 \\
Replicas average & 2.68 & 37.72 \\
Outer parameters & \textbf{2.67} & \textbf{37.78} \\
\bottomrule
\end{tabular}
%}
\caption{\textbf{Which parameters to evaluate?}: Evaluating the outer parameters, where each fragment has been synchronized at a different moment in time, yields better performance than any inner parameters.}
\label{tab:evaluate_mode}
\end{table}

\begin{figure*}
\centering
 \includegraphics[width=.99\linewidth]{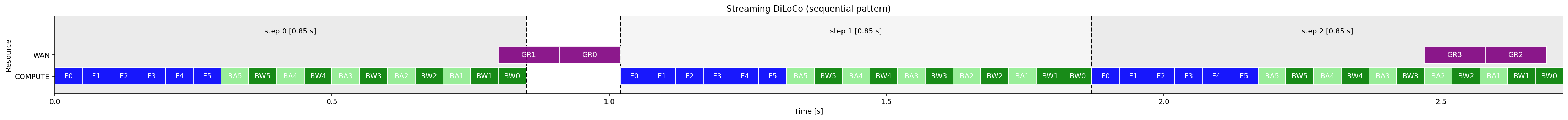}\\
 \includegraphics[width=.99\linewidth]{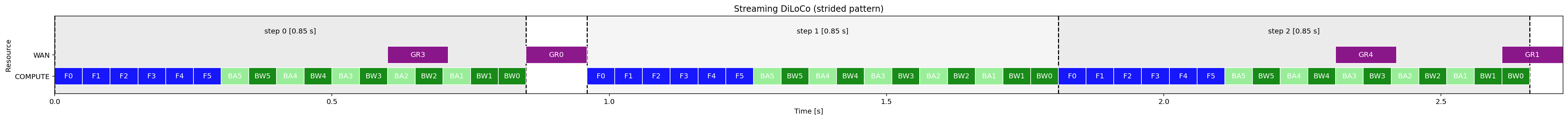} \\
\caption{Simulation of a schedule interleaving forward passes (in \textcolor{blue}{blue}), backward passes w.r.t. activations and weights (resp. in \textcolor{green}{light} and \textcolor{teal}{dark green}), and (outer) gradient reduction (in \textcolor{violet}{purple}) for Streaming DiLoCo, respectively with a sequential and strided pattern.}
\label{fig:schedule_strided}
\end{figure*}

\paragraph{Sequential vs strided patterns.} The choice of the synchronization pattern (\autoref{fig:streaming_pattern}), has a slight impact on the ML performance (\autoref{fig:fragment_size_loss}) but also on the compute utilization (\autoref{fig:bandwidth_100b_stride}). Indeed, as better seen in \autoref{fig:schedule_strided}, the strided pattern will never have multiple early layers to be synchronize together. Therefore, it is easier to overlap their communication with the first few layers' forward of the next step.

\paragraph{Compute utilization.} We report in \autoref{tab:simulation_hps} the amount of Gbit/s required, per method, to reach a certain level of compute utilization. See \autoref{fig:bandwdith} for a figure view of this table. For DiLoCo and Streaming DiLoco (and variants thereof), we use $H=100$ inner steps. For Streaming DiLoCo (and variants thereof), we use a fixed fragment size of 3 layers; therefore, deeper networks have more fragments: for 1B, 10B, and 100B model scales, it is respectively 8, 16, and 36 fragments. Also, respectively per model scales, a fragment is synchronized every 11, 5, and 2 steps. While the synchronization seems to be more frequent for deeper networks,  from the perspective of particular fragment, it is synchronized roughly every $H=100$ steps. To estimate the compute utilization in \autoref{tab:simulation_hps} and \autoref{fig:bandwdith}, the time spent per step doing computation (forward \& backward) is critical: we report respectively 0.1s, 0.8s, and 4.9s based on each model scale flops profile, a reasonable amount of chips, and a MFU of $60\%$.

\begin{table*}[t]
\centering
\resizebox{1.0\linewidth}{!}{%
%\vspace*{-0.3cm}
\begin{tabular}{@{}ccc|c|ccccc@{}}
\toprule
\multirow{2}{*}{Model size} & \multirow{2}{*}{\# layers} & \multirow{2}{*}{Step time} &   \multirow{2}{*}{Method} & \multicolumn{5}{c}{Gbit/s to reach a compute utilization $\texttt{CU} = $?} \\
&&&&$50\%$ &  $80\%$ &  $90\%$ &  $95\%$ &  $99\%$  \\
\midrule
\multirow{5}{*}{1B} & \multirow{5}{*}{24} & \multirow{5}{*}{0.1s} & Data-Parallel & 86.8 & 152.6 & 184.2 & 222.3 & 569.0 \\
& & & Vanilla DiLoCo & 1.4 & 6.2 & 13.3 & 23.3 & 86.8 \\
& & & Streaming DiLoCo & 1.4 & 5.2 & 9.1 & 16.0 & 28.1 \\
& & & Streaming DiLoCo w/ overlapped com. & 1.4 & 4.3 & 6.2 & 9.1 & 11.0 \\
& & & Streaming DiLoCo w/ overlapped FP4 com. & 0.4 & 0.9 & 1.7 & 2.0 & 3.0 \\
\midrule
\multirow{5}{*}{10B} & \multirow{5}{*}{48} & \multirow{5}{*}{0.8s} & Data-Parallel & 104.8 & 222.3 & 222.3 & 268.3 & 471.5 \\
& & & Vanilla DiLoCo & 1.7 & 7.5 & 16.0 & 33.9 & 104.8 \\
& & & Streaming DiLoCo & 1.7 & 5.2 & 9.1 & 13.3 & 19.3 \\
& & & Streaming DiLoCo w/ overlapped com. & 1.7 & 3.6 & 5.2 & 6.2 & 7.5 \\
& & & Streaming DiLoCo w/ overlapped FP4 com. & 0.4 & 0.9 & 1.4 & 1.4 & 1.7 \\
\midrule
\multirow{5}{*}{100B} & \multirow{5}{*}{108} & \multirow{5}{*}{4.9s} & Data-Parallel & 184.2 & 323.8 & 390.7 & 390.7 & 471.5 \\
& & & Vanilla DiLoCo & 3.0 & 11.0 & 23.3 & 49.4 & 184.2 \\
& & & Streaming DiLoCo & 2.4 & 6.2 & 9.1 & 11.0 & 19.3 \\
& & & Streaming DiLoCo w/ overlapped com. & 1.7 & 3.6 & 4.3 & 5.2 & 5.2 \\
& & & Streaming DiLoCo w/ overlapped FP4 com. & 0.5 & 0.9 & 1.1 & 1.1 & 1.4 \\
\bottomrule
\end{tabular}
}
\caption{\textbf{Simulation}: we estimate the step time (pure compute) of 10B and 100B based on the required flops using \cite{kaplan2020scalinglawsneurallanguage} rule and using a MFU of 60\%. For all DiLoCo and Streaming DiLoCo-variants, we use $H=100$. For all Streaming DiLoCo-variants, we use a fragment size of 3 layers.}
\label{tab:simulation_hps}
\end{table*}

\begin{figure*}[t]
\centering
\captionsetup[subfigure]{justification=centering}
\begin{subfigure}{0.3\linewidth}
  \centering
  \includegraphics[width=1\linewidth]{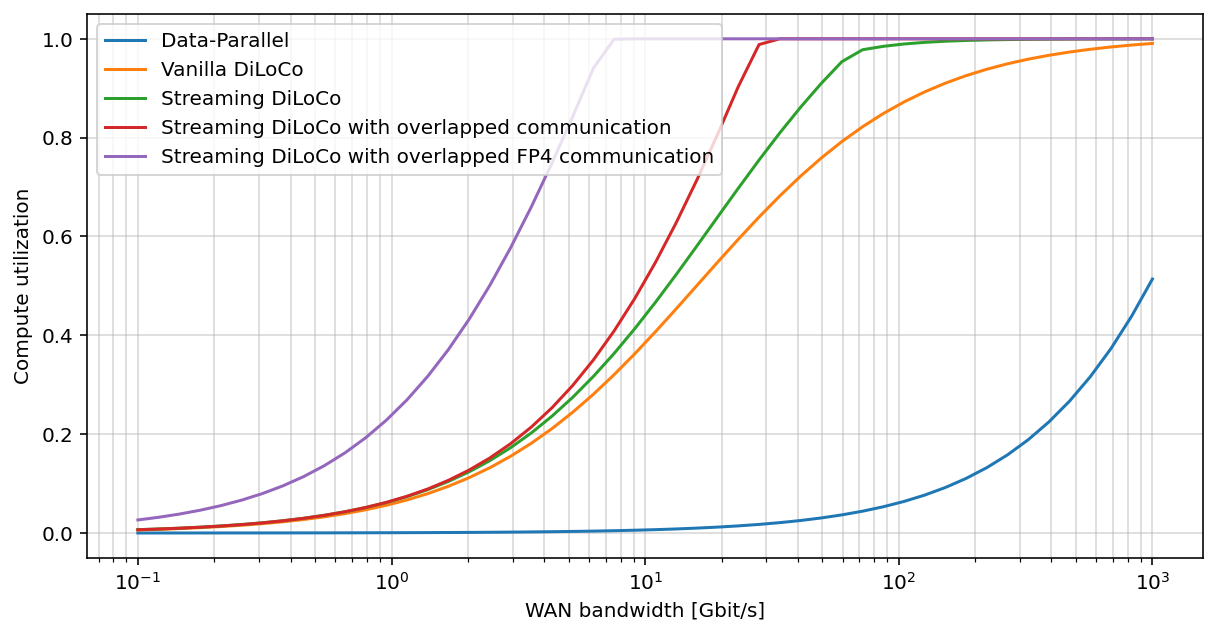}
  \caption{1s step time}
  \label{fig:bandwdith_100b_1s}
\end{subfigure}
\begin{subfigure}{0.3\linewidth}
  \centering
  \includegraphics[width=1\linewidth]{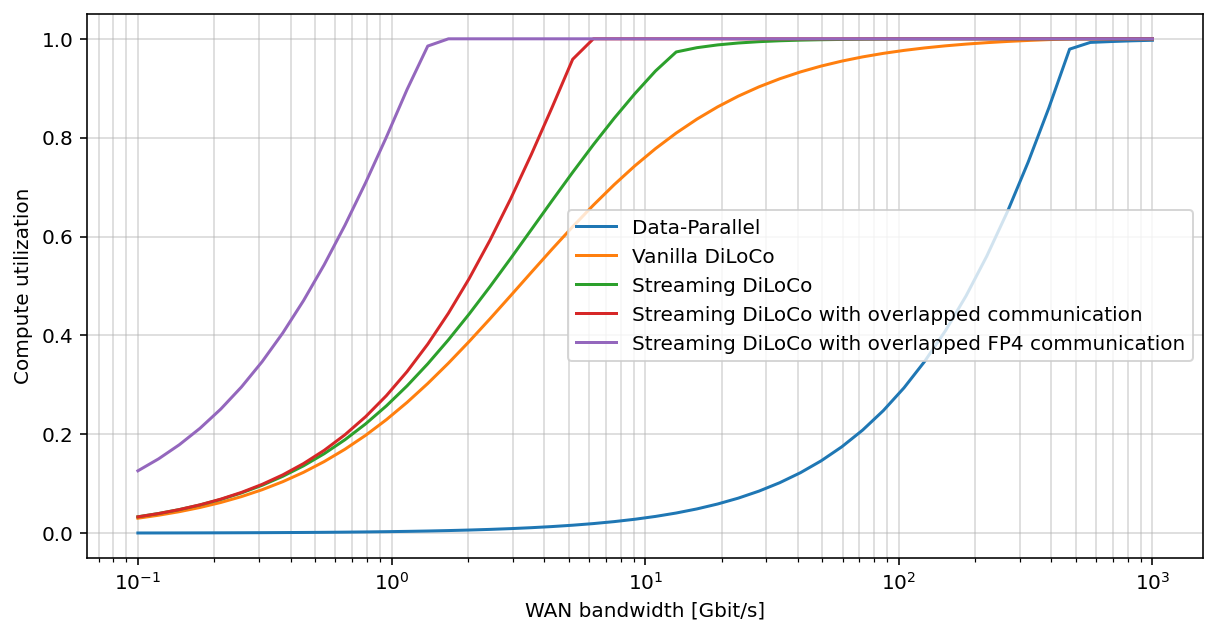}
  \caption{5s step time}
  \label{fig:bandwdith_100b_5s}
\end{subfigure}
\begin{subfigure}{0.3\linewidth}
  \centering
  \includegraphics[width=1\linewidth]{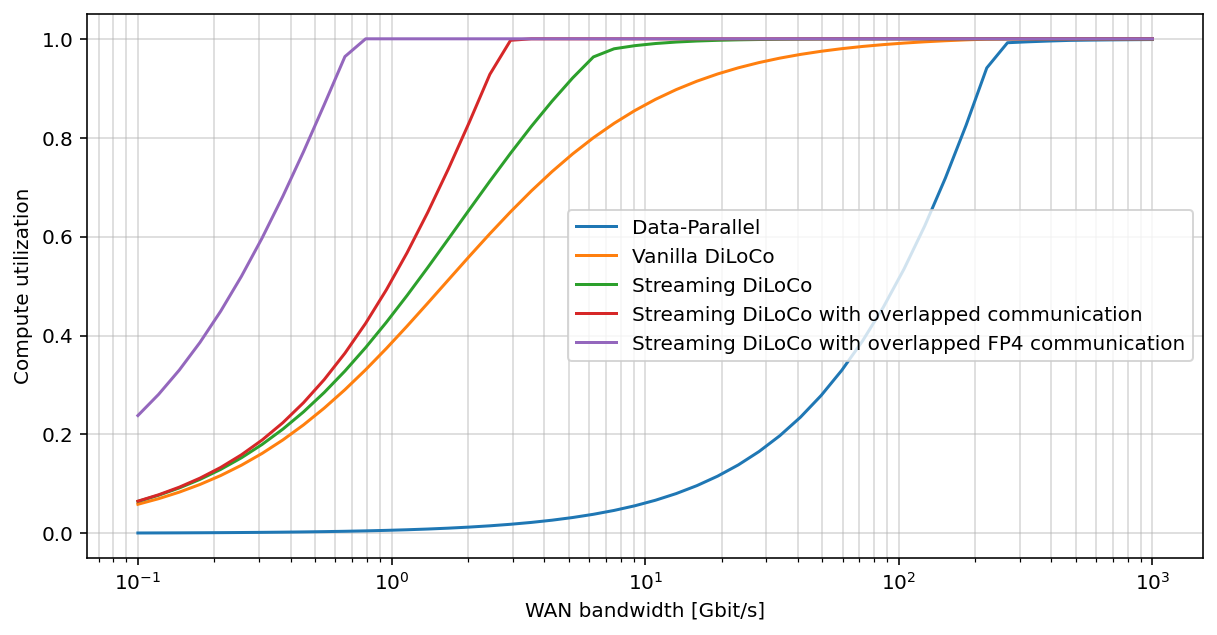}
  \caption{10s step time}
  \label{fig:bandwdith_100b_10s}
\end{subfigure}
\caption{\textbf{Compute Utilization} for a 100 billion parameters when the step time (pure compute) is 1 second, 5 seconds, and 10 seconds.}
\label{fig:bandwdith_sec}
\end{figure*}

\paragraph{Compute utilization with various speeds.} Varying the time spent per step to do pure computation (forward \& backward) affects the compute utilization: e.g. for a fixed bandwidth and thus fixed communication time, longer step time, will improve compute utilization. We report in \autoref{fig:bandwdith_sec}, simulated compute utilization when using, at 100B model scale, a compute step time of 1 second, 5 seconds, and 10 seconds. 

\begin{figure*}[t]
\centering
\captionsetup[subfigure]{justification=centering}
\begin{subfigure}{0.49\linewidth}
  \centering
  \includegraphics[width=1\linewidth]{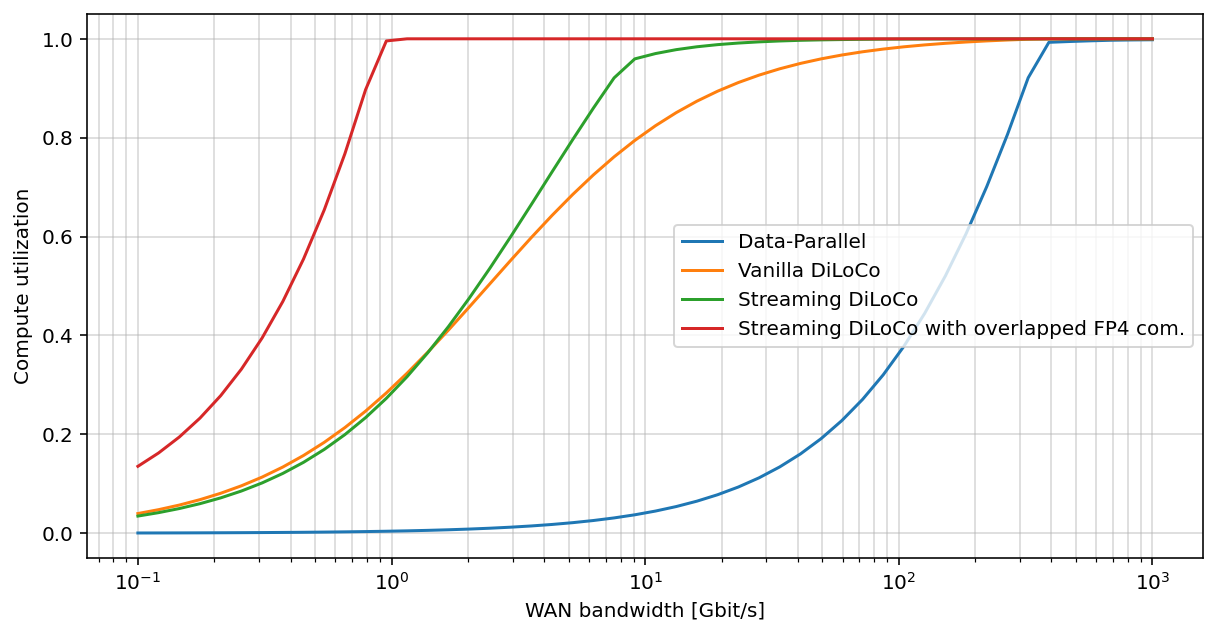}
  \caption{Llama405B.}
  \label{fig:bandwdith_llama}
\end{subfigure}\hfill
\begin{subfigure}{0.49\linewidth}
  \centering
  \includegraphics[width=1\linewidth]{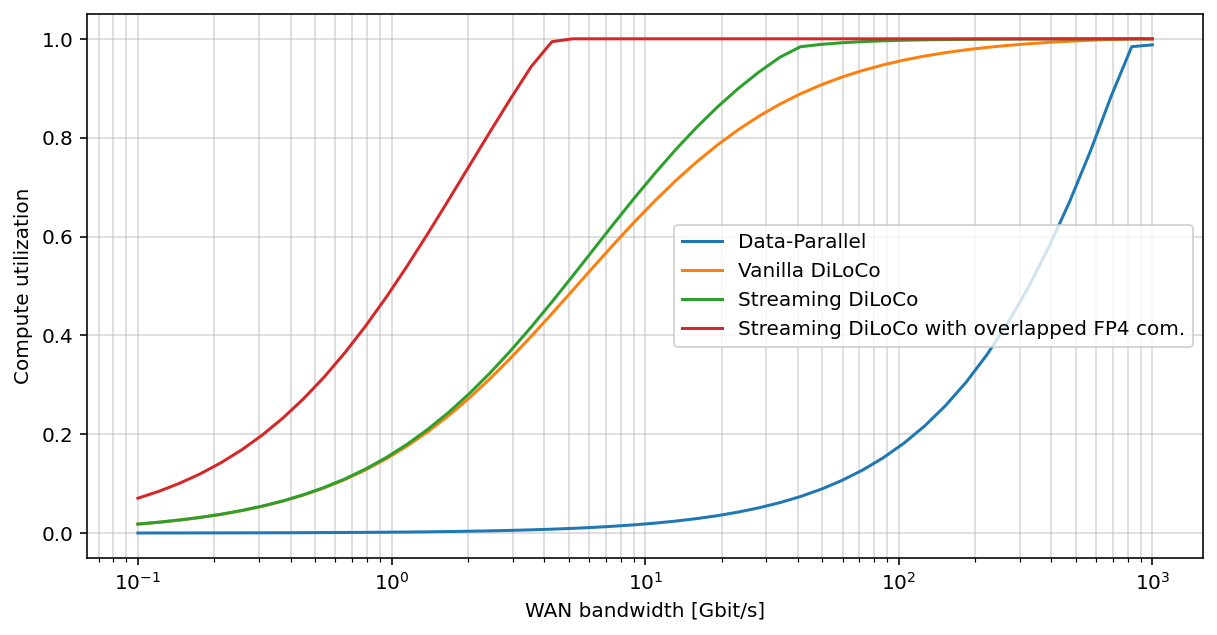}
  \caption{DeepSeek-V3 (671B total, 35B activated).}
  \label{fig:bandwdith_deepseek}
\end{subfigure}\hfill
\caption{\textbf{Compute Utilization} simulated across a range of bandwidth for Llama405 and DeepSeek-V3, using step time estimated from respective papers.}
\label{fig:bandwdith_external}
\end{figure*}

\paragraph{Compute Utilization on Llama and DeepSeek.} We estimate in \autoref{fig:bandwdith_external}, the compute utilization of our method vs baselines on top of Llama405 \citep{grattafiori2024llama3herdmodels} and DeepSeek-V3 \citep{deepseekai2024deepseekv3technicalreport}. For each, we estimate their step time from the respective paper: 26.9 seconds for Llama (first stage of pretraining) and 20.1 seconds for DeepSeek, using the most charitable estimation everytime. Notably, for DeepSeek-V3 (\autoref{fig:bandwdith_deepseek}), only 35 billion parameters are activated per token due to their MoE architecture \citep{shazeer2017outrageouslylargeneuralnetworks}. However, the total 671 billion parameters are synchronized between replicas, massively increasing the amount of bits to transfer. In that case, in our simulation, our method (in \textcolor{BrickRed}{red}) can be close to 100\% compute utilization with 4 Gbits per second vs 1 Tbit per second for Data-Parallel.

\paragraph{Scaling performance.} We report in \autoref{tab:scaling}, the evaluation loss on C4 and accuracy on HellaSwag \citep{zellers2019hellaswagmachinereallyfinish}, Piqa \citep{bisk2019piqareasoningphysicalcommonsense}, and Arc-Easy \citep{clark2018arc}, for four different methods across 6 model scales. See \autoref{sec:scaling} for the initial discussion. Performance across scales are roughly similar among all considered methods, with usually a slight advantage for Data-Parallel. We found in practice this advantage to disappear when doing a more realistic overtraining with larger token budget in \autoref{sec:overtraining}.

\begin{table*}[t]
\centering
\resizebox{1.0\linewidth}{!}{%
%\vspace*{-0.3cm}
\begin{tabular}{@{}cc|ccccc@{}}
\toprule
Model size & Flops & Method & Eval Loss $\downarrow$ & HellaSwag $\uparrow$ & Piqa $\uparrow$ & Arc Easy $\uparrow$ \\
\midrule
\multirow{4}{*}{35M} & \multirow{4}{*}{1.5e17} & Data-Parallel & 3.51 & 24.62 & 57.89 & 29.65 \\
& &  DiLoCo H=30 & 3.54 & 24.53 & 58.11 & 29.65 \\
& &  Streaming DiLoCo with overlapped FP4 com., H=30 & 3.53 & 24.46 & 57.67 & 30.53 \\
& &  Streaming DiLoCo with overlapped FP4 com., H=100 & 3.56 & 24.80 & 57.89 & 29.12 \\
\midrule
\multirow{4}{*}{100M} & \multirow{4}{*}{9.4e17} & Data-Parallel & 3.19 & 26.94 & 60.12 & 30.35 \\
& &  DiLoCo H=30 & 3.21 & 26.59 & 60.50 & 29.12 \\
& &  Streaming DiLoCo with overlapped FP4 com., H=30 & 3.21 & 26.97 & 59.58 & 31.40 \\
& &  Streaming DiLoCo with overlapped FP4 com., H=100 & 3.22 & 26.68 & 60.39 & 31.93 \\
\midrule
\multirow{4}{*}{200M} & \multirow{4}{*}{4e18} & Data-Parallel & 2.97 & 29.86 & 63.71 & 35.44 \\
& &  DiLoCo H=30 & 2.98 & 29.71 & 62.30 & 33.68 \\
& &  Streaming DiLoCo with overlapped FP4 com., H=30 & 2.98 & 29.67 & 61.92 & 34.39 \\
& &  Streaming DiLoCo with overlapped FP4 com., H=100 & 3.00 & 29.27 & 62.13 & 34.21 \\
\midrule
\multirow{4}{*}{300M} & \multirow{4}{*}{1.4e19} & Data-Parallel & 2.80 & 33.46 & 64.69 & 34.91 \\
& &  DiLoCo H=30 & 2.81 & 33.87 & 64.74 & 34.74 \\
& &  Streaming DiLoCo with overlapped FP4 com., H=30 & 2.81 & 33.66 & 63.49 & 35.09 \\
& &  Streaming DiLoCo with overlapped FP4 com., H=100 & 2.83 & 33.00 & 63.71 & 34.39 \\
\midrule
\multirow{4}{*}{500M} & \multirow{4}{*}{4.7e19} & Data-Parallel & 2.67 & 38.68 & 66.49 & 37.19 \\
& &  DiLoCo H=30 & 2.68 & 38.37 & 65.61 & 36.32 \\
& &  Streaming DiLoCo with overlapped FP4 com., H=30 & 2.67 & 38.10 & 66.21 & 34.91 \\
& &  Streaming DiLoCo with overlapped FP4 com., H=100 & 2.69 & 37.40 & 65.51 & 34.74 \\
\midrule
\multirow{4}{*}{1B} & \multirow{4}{*}{1.9e20} & Data-Parallel & 2.49 & 46.60 & 68.93 & 39.65 \\
& &  DiLoCo H=30 & 2.49 & 46.56 & 68.82 & 36.84 \\
& &  Streaming DiLoCo with overlapped FP4 com., H=30 & 2.48 & 46.60 & 69.04 & 39.12 \\
& &  Streaming DiLoCo with overlapped FP4 com., H=100 & 2.50 & 46.00 & 68.82 & 38.42 \\
\midrule
\multirow{4}{*}{4B} & \multirow{4}{*}{2e21} & Data-Parallel & 2.25 & 59.56 & 72.42 & 43.51 \\
& &  DiLoCo H=30 & - & - & - & - \\
& &  Streaming DiLoCo with overlapped FP4 com., H=30 & - & - & - & - \\
& &  Streaming DiLoCo with overlapped FP4 com., H=100 & 2.26 & 59.02 & 72.52 & 43.16 \\
\bottomrule
\end{tabular}
}
\caption{\textbf{Scaling} from 35 million parameters to 4 billion parameters using a chinchilla-optimal number of flops/tokens. We train on the C4 dataset, and report the evaluation loss on its validation set.}
\label{tab:scaling}
\end{table*}

\paragraph{Scaling with variable number of replicas.} Contrarely to Data-Parallel, changing the number of replicas for DiLoCo is not mathematically equivalent due to the local training, happening independentely for each replicas. We display in \autoref{tab:scaling_m4}, a scaling from 35 million parameters to 1 billion parameters on the C4 dataset of our method, Streaming DiLoCo with overlapped FP4 communication, with different number of replicas $M=\{2,4\}$ and different frequencies of synchronization $H=\{30, 100\}$. Likewise, in \autoref{tab:overtrain_steps_m4}, we showcase token budget overtraining at 1 billion parameters on the Dolma dataset.

\begin{table*}[t]
\centering
%\resizebox{1.0\linewidth}{!}{%
%\vspace*{-0.3cm}
\begin{tabular}{@{}cc|cccccc@{}}
\toprule
Model size & Flops & $M$ & $H$ & Eval Loss $\downarrow$ & HellaSwag $\uparrow$ & Piqa $\uparrow$ & Arc Easy $\uparrow$ \\
\midrule
\multirow{4}{*}{35M} & \multirow{4}{*}{1.5e17} & 2 & 30 & 3.53 & 24.46 & 57.67 & 30.53 \\
& &  4 & 30 & 3.60 & 24.50 & 56.09 & 28.60 \\
& &  2 & 100 & 3.56 & 24.80 & 57.89 & 29.12 \\
& &  4 & 100 & 3.64 & 24.67 & 56.75 & 26.84 \\
\midrule
\multirow{4}{*}{100M} & \multirow{4}{*}{9.4e17} & 2 & 30 & 3.21 & 26.97 & 59.58 & 31.40 \\
& &  4 & 30 & 3.25 & 26.24 & 59.74 & 32.63 \\
& &  2 & 100 & 3.22 & 26.68 & 60.39 & 31.93 \\
& &  4 & 100 & 3.29 & 26.54 & 60.34 & 29.82 \\
\midrule
\multirow{4}{*}{200M} & \multirow{4}{*}{4e18} & 2 & 30 & 2.98 & 29.67 & 61.92 & 34.39 \\
& &  4 & 30 & 3.02 & 29.09 & 62.89 & 35.44 \\
& &  2 & 100 & 3.00 & 29.27 & 62.13 & 34.21 \\
& &  4 & 100 & 3.05 & 28.53 & 61.10 & 33.51 \\
\midrule
\multirow{4}{*}{300M} & \multirow{4}{*}{1.4e19} & 2 & 30 & 2.81 & 33.66 & 63.49 & 35.09 \\
& &  4 & 30 & 2.84 & 32.54 & 64.42 & 34.74 \\
& &  2 & 100 & 2.83 & 33.00 & 63.71 & 34.39 \\
& &  4 & 100 & 2.87 & 32.02 & 64.25 & 35.44 \\
\midrule
\multirow{4}{*}{500M} & \multirow{4}{*}{4.7e19} & 2 & 30 & 2.67 & 38.10 & 66.21 & 34.91 \\
& &  4 & 30 & 2.70 & 36.95 & 65.72 & 35.26 \\
& &  2 & 100 & 2.69 & 37.40 & 65.51 & 34.74 \\
& &  4 & 100 & 2.73 & 36.02 & 66.27 & 35.09 \\
\midrule
\multirow{4}{*}{1B} & \multirow{4}{*}{1.9e20} & 2 & 30 & 2.48 & 46.60 & 69.04 & 39.12 \\
& &  4 & 30 & 2.50 & 45.25 & 67.95 & 39.12 \\
& &  2 & 100 & 2.50 & 46.00 & 68.82 & 38.42 \\
& &  4 & 100 & 2.53 & 44.74 & 68.34 & 38.25 \\
\bottomrule
\end{tabular}
%}
\caption{\textbf{Scaling} from 35 million parameters to 1 billion parameters Streaming DiLoCo with overlapped FP4 communication and with two different synchronization frequencies $H=\{30, 100\}$ and number of DiLoCo replicas $M=\{2, 4\}.$}
\label{tab:scaling_m4}
\end{table*}

\begin{table*}[t]
\centering
\resizebox{1.0\linewidth}{!}{%
%\vspace*{-0.3cm}
\begin{tabular}{@{}l|cc|cccc@{}}
\toprule
Method & Token Budget & Terabytes exchanged $\downarrow$ & Eval Loss $\downarrow$ & HellaSwag $\uparrow$ & Piqa $\uparrow$ & Arc Easy $\uparrow$ \\
\midrule
\multirow{3}{*}{Data-Parallel} & 25B & 441 & 2.67 & \textbf{42.09} & 67.35 & \textbf{40.42} \\
 & 100B & 1,767 & 2.52 & 49.78 & 69.15 & \textbf{44.03} \\
 & 250B & 4,418 & \textbf{2.45} & 53.86 & 70.45 & \textbf{44.21} \\
\midrule
\multirow{3}{*}{Our method, M=2} & 25B & 1.10 & \textbf{2.66} & 42.08 & \textbf{67.46} & 38.42 \\ 
& 100B & 4.42 & \textbf{2.51} & \textbf{49.98} & \textbf{69.96} & \textbf{44.03} \\
& 250B & 11.05 & \textbf{2.45} & \textbf{54.24} & \textbf{71.38} & 41.92 \\
\midrule
\multirow{3}{*}{Our method, M=4} & 25B & 0.55 & 2.73 & 38.93 & 66.92 & 39.64 \\ 
& 100B & 2.21 & 2.54 & 48.35 & 69.42 & 40.52 \\
& 250B & 5.52 & 2.47 & 52.20 & 70.29 & 42.45 \\

\bottomrule
\end{tabular}
}
\caption{\textbf{Overtraining} on the Dolma dataset with a 1 billion parameters model, and with an increasing token budgets (25B, 100B, and 250B). We report here for our model both with $M=2$ and $M=4$ DiLoCo replicas. With twice more replicas, the global batch size is doubled, and twice less steps are done. It is also thus roughly twice faster, but come with slightly worse performance. Our method is the final model: Streaming DiLoCo with overlapped FP4 communication.}
\label{tab:overtrain_steps_m4}
\end{table*}

\begin{figure*}[t]
\captionsetup[subfigure]{justification=centering}
\begin{subfigure}{0.5\linewidth}
  \centering
  \includegraphics[width=1\linewidth]{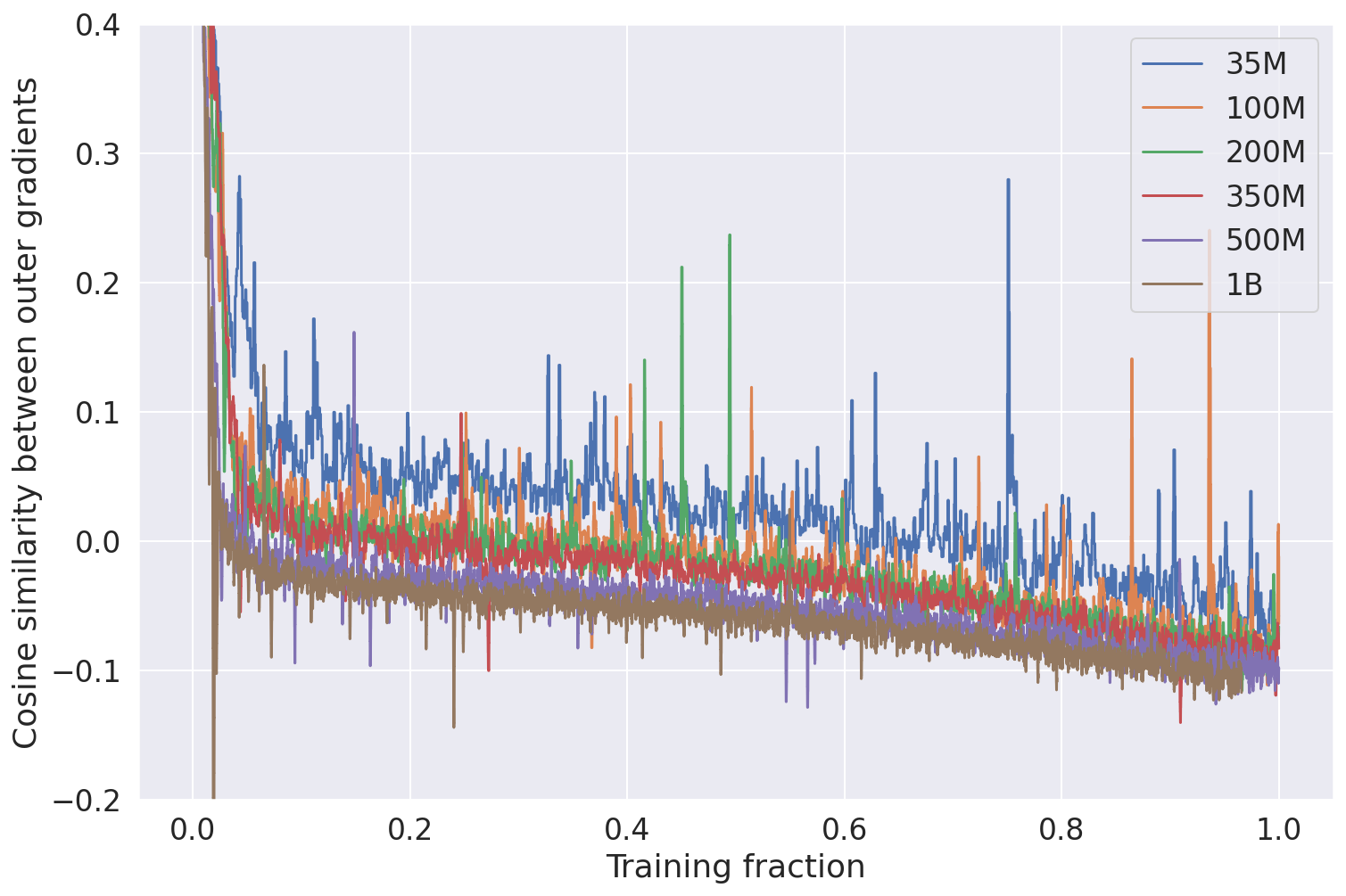}
  \caption{All fragments but the embedding}
  \label{fig:cosine_all}
\end{subfigure}
\begin{subfigure}{0.5\linewidth}
  \centering
  \includegraphics[width=1\linewidth]{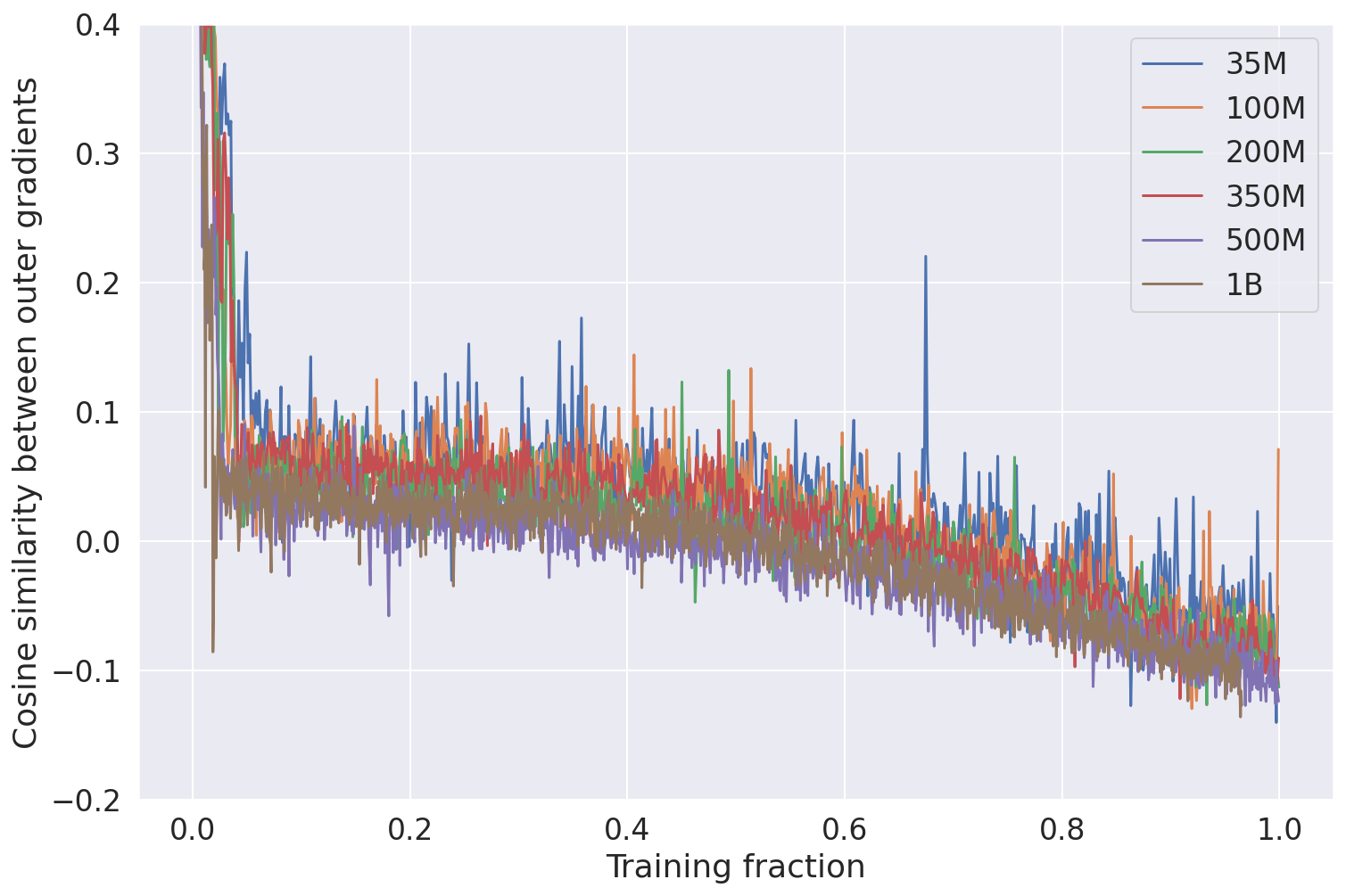}
  \caption{Embedding fragment}
  \label{fig:cosine_emb}
\end{subfigure}
\caption{\textbf{Cosine similarity between the outer gradients} across scales.}
\label{fig:cosine}
\end{figure*}

\paragraph{Outer gradients' cosine similarity.} We observe in \autoref{fig:cosine} the cosine similarity per scale between each replica's outer gradients for respectively all parameters but the embeddings (\autoref{fig:cosine_all}) and only the embeddings (\autoref{fig:cosine_emb}). For both, the cosine similarity starts from slightly correlated ($\approx 0.1$), spends of the training time to be close to orthogonal ($\approx 0.0$), and ends slightly inversely correlated ($\approx -0.1$) as we reach the fluctuation phase. Note also that the larger the model size, the lower is overall the cosine similarity. 

\begin{figure*}[t]
\centering
  \includegraphics[width=1\linewidth]{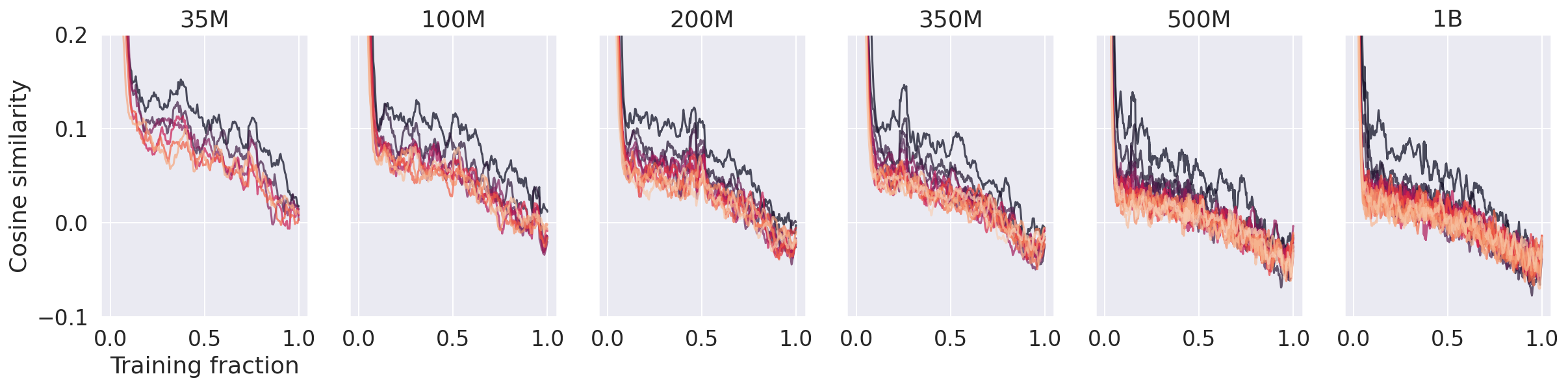}
\caption{\textbf{Cosine similarity between the outer gradients} across scales. Each line is a transformer layer, with darker colors being earlier layers and lighter colors later layers.}
\label{fig:cosine_scale}
\end{figure*}

We also plot in \autoref{fig:cosine_scale} the cosine similarity per scale and per transformer layer. Notably, the first transformer layer at each scale has a significantly higher similarity, at every model scales.

\end{document}